\documentclass[11pt, a4paper, logo, copyright]{googledeepmind}

\usepackage[super,numbers]{natbib}
\usepackage{authblk}

\usepackage{amsmath,amsfonts,bm}

\def\eqref#1{equation~\ref{#1}}

\def\1{\bm{1}}

\DeclareMathAlphabet{\mathsfit}{\encodingdefault}{\sfdefault}{m}{sl}
\SetMathAlphabet{\mathsfit}{bold}{\encodingdefault}{\sfdefault}{bx}{n}

\usepackage{times}
\usepackage{latexsym}
\usepackage[T1]{fontenc}
\usepackage[utf8]{inputenc}
\usepackage{microtype}
\usepackage{inconsolata}
\usepackage{graphicx}
\usepackage{balance}       
\usepackage{graphics}      
\usepackage{hyperref}
\usepackage{nameref}
\usepackage{color}
\usepackage{booktabs}
\usepackage{textcomp}
\usepackage{subcaption}
\usepackage{enumerate}
\usepackage{xcolor}
\usepackage{lipsum}
\usepackage{makecell}
\usepackage{multicol}
\usepackage{multirow}
\usepackage{array}
\usepackage{verbatimbox}
\usepackage{enumitem}
\usepackage{amsmath}
\usepackage{float}
\usepackage{stfloats}
\usepackage{graphicx}
\usepackage{amsthm}
\usepackage{listings}
\usepackage{caption} 
\usepackage[export]{adjustbox}
\usepackage{xspace}
\usepackage{epsfig}
\usepackage[linesnumbered]{algorithm2e}
\usepackage{algpseudocode}
\usepackage{tabularx}
\usepackage{arydshln}
\usepackage[bottom]{footmisc}
\usepackage{tcolorbox}
\usepackage{stackengine}
\usepackage{placeins}
\usepackage[normalem]{ulem}
\useunder{\uline}{\ul}{}
\tcbuselibrary{breakable}

\definecolor{E+F}{RGB}{	255, 99, 71}
\definecolor{B+F}{RGB}{255, 165, 0}
\definecolor{E+I}{RGB}{	173, 216, 230}
\definecolor{B+I}{RGB}{	30, 144, 255}
\definecolor{D}{RGB}{	60, 179, 113}

\newcommand{\methods}{\textit{Methods }}
\newcommand{\results}{\textit{Results }}

\definecolor{maroon}{cmyk}{0,0.87,0.68,0.32}

\usepackage{tikz}

\newcommand{\revisiontwo}[1]{\textcolor{black}{#1}}

\definecolor{darkgreen}{rgb}{0.0, 0.5, 0.0}
\definecolor{usercolor}{RGB}{200, 230, 250} 
\definecolor{coachcolor}{RGB}{180, 250, 180} 

\title{A Scalable Framework for Evaluating Health Language Models}
\reportnumber{} 
\renewcommand{\today}{2026-01-28}

\author[1,$\dagger$, *]{Neil Mallinar}
\author[1,$\dagger$]{A. Ali Heydari}
\author[1]{Xin Liu}
\author[1]{Anthony Z. Faranesh}
\author[1]{Brent Winslow}
\author[1]{Nova Hammerquist}
\author[2]{Benjamin Graef}
\author[1, *]{Cathy Speed}
\author[1]{Mark Malhotra}
\author[1]{Shwetak Patel}
\author[1, *]{Javier L. Prieto}
\author[1, $\ddagger$]{Daniel McDuff}
\author[1, $\ddagger$]{Ahmed A. Metwally}
\affil[1]{Google Research,}
\affil[2]{Vituity\protect\\}
\affil[$\dagger$]{Equal contribution,}

\affil[$\ddagger$]{Corresponding authors: \texttt{\{aametwally, dmcduff\}}@google.com.}

\begin{abstract}

Large language models (LLMs) have emerged as powerful tools for analyzing and interpreting complex datasets. Recent studies demonstrate their potential to generate useful, personalized responses when provided with patient-specific health information that encompasses lifestyle, biomarkers, and context. As LLM-driven health applications are increasingly adopted, rigorous and efficient one-sided evaluation methodologies are crucial to ensure response quality across multiple dimensions, including accuracy, personalization, relevance and safety. However, current evaluation practices, particularly for open-ended text responses, heavily rely on human experts. This approach introduces human factors (perspectives, potential biases, inconsistencies) and is often cost-prohibitive, labor-intensive, and hinders scalability, especially in complex domains like healthcare where response assessment necessitates domain expertise and considers multifaceted patient data, which is often nuanced and diverse. In this work, we introduce Adaptive Precise Boolean rubrics: an evaluation framework that aims to streamline human and automated evaluation of open-ended questions by identifying critical gaps in model responses using a minimal set of targeted rubrics questions. Our approach is based on recent work in more general evaluation settings that contrasts a smaller set of complex evaluation targets with a larger set of more precise, granular targets answerable with simple boolean responses. We validate this approach in metabolic health, a domain encompassing diabetes, cardiovascular disease, and obesity. Our results demonstrate that Adaptive Precise Boolean rubrics yield  substantially higher inter-rater agreement among both expert and non-expert human evaluators, as well as in automated assessments, compared to traditional Likert scales, while requiring approximately half the evaluation time of Likert-based methods. This enhanced efficiency and scalability, particularly through automated evaluation and non-expert contributions, paves the way for more extensive and cost-effective evaluation of LLMs in health. 

\end{abstract}
\begin{document}
\let\thefootnote\relax\footnotetext{$^*$Work done while at Google Research.}
\maketitle
\section*{Introduction}
\label{sec:intro}
\phantomsection

The convergence of artificial intelligence and healthcare is being fundamentally reshaped by the use of language models (LMs) that can perform complex linguistic operations ~\citep{Arora2023-fl}. These models, capable of processing and reasoning multimodal health data, are not merely tools, but represent a potential paradigm shift in how medical knowledge and information is accessed, synthesized, and utilized, with applications such as medical question and answering ~\citep{Singhal2022-dp}, differential diagnosis \citep{McDuff2023-rc}, electronic health record reasoning \citep{Shi2024-va}, and biomedical research discovery  \citep{Gottweis2025-kr}. This potential of LLMs extends beyond traditional healthcare, with accelerating adoption observed across a wide spectrum of consumer health applications including sleep and fitness coaching \citep{Cosentino2024-fs}, wearable insights extraction \citep{Merrill2024-ad}, personalized health coaching \citep{heydari2025}, symptom checkers \citep{Fraser2023-ga}, and clinical examination dialogues \citep{Tu2024-mr}.

\begin{figure*}
    \centering
    \includegraphics[width=0.95\textwidth]{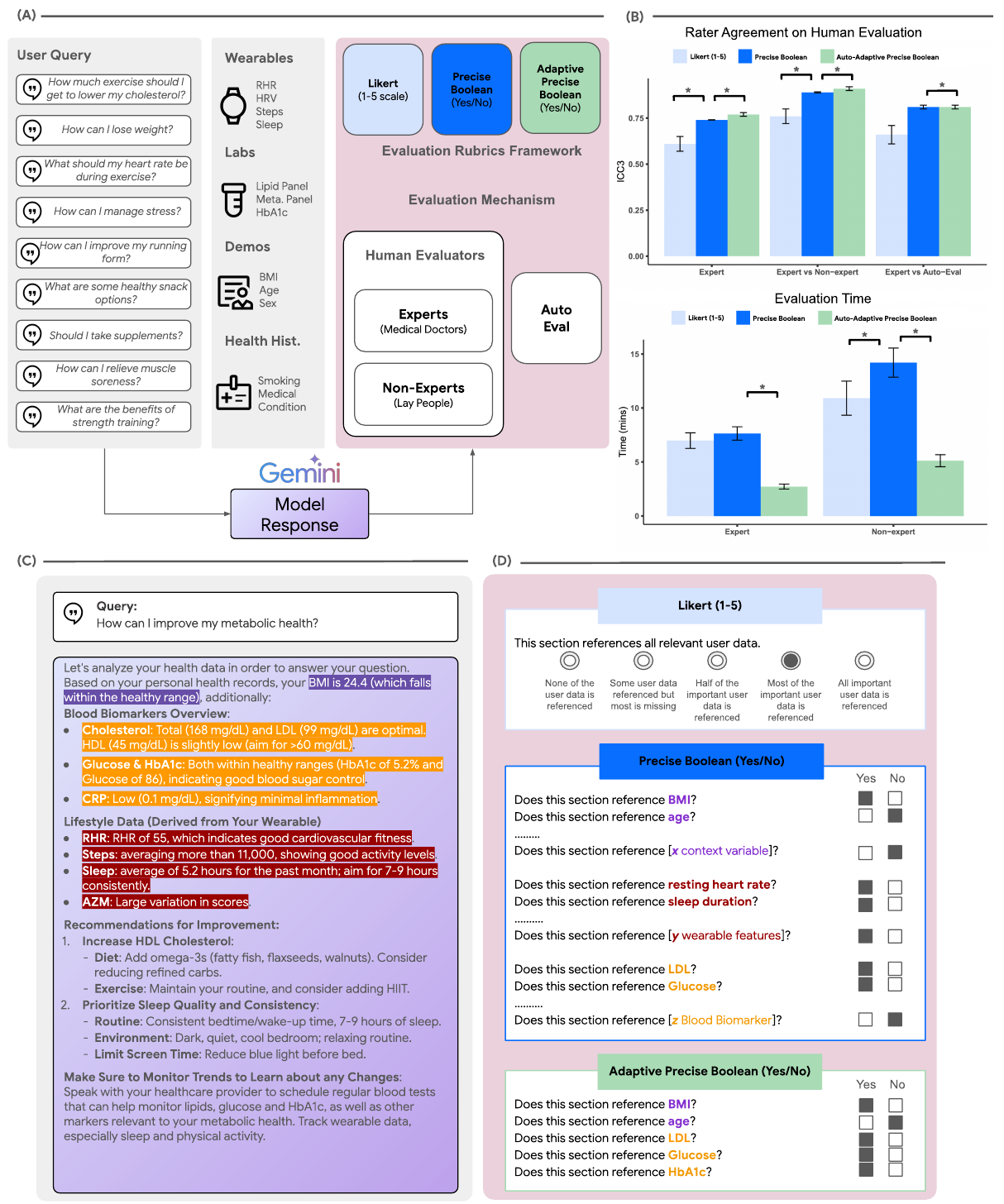}
    \caption{\textbf{A precision rubric for evaluating open-ended language responses to health queries.} (A) A set of representative health queries and wearable data are used to construct inputs to the language model, which are then evaluated using our proposed evaluation rubric. (B) Our Results indicate that the proposed Precise Boolean rubric leads to higher agreement with humans (both experts and non-expert). Moreover, leveraging our proposed Adaptive Precise Boolean, the time being used in evaluation is halved compared to the traditional Likert scale, while maintaining high rater agreement and high quality evaluation. (C) An example of a query and response highlighting references to specific relevant parts of the response. (D) Examples of evaluation rubrics for assessing the generated response to input query}
    \label{fig:figure1}
\end{figure*}

Evaluation is a critical component in the development and validation of AI models. In safety-critical applications such as healthcare, this process must extend beyond standard benchmarks to assess model robustness. Techniques such as adversarial testing and red-teaming are essential to systematically probe for vulnerabilities and elicit unsafe or factually incorrect responses not typically observed in curated datasets \citep{adversarial-robustness2, adversarial-robustness1}. However, these rigorous evaluations often rely on manual, creative human effort, which compounds the challenge of scalability and highlights the need for more efficient methodologies \citep{evaluating-llms-generalization}. Unlike discriminative tasks (e.g., multiple choice questions/classification), open-ended generative language tasks are more complex to evaluate as there can be manifold answers with multiple properties (e.g., accuracy, comprehensiveness, relevance) - there is not a single ``right answer" \citep{Liang2022-lh}. Questionnaires tend to fall into four categories: \emph{yes} or \emph{no} questions (binary choice), multiple choice (multichoice), open-ended text responses, and responses on ordinal scales of approval or disapproval \citep{Likert1932-hp}. The last of these can be used to capture additional nuance where there are degrees of “correctness” and no single correct category. Indeed, previous studies in the health domain have used Likert scales to grade open-ended LLM responses \citep{Singhal2022-dp,McDuff2023-rc,Cosentino2024-fs}, however, it was observed that rater agreement measures in such cases can be suboptimal. Where an objective correct answer is not present, it is necessary to assess the quality of a scoring rubric by measuring the agreement of qualified raters \citep{Landis1977-fa}. Identifying evaluation systems that improve on such agreement measures is of broad interest in the evaluation literature.

Comprehensive evaluation of LLM-based systems requires a robust evaluation approach, encompassing a spectrum of methodologies ranging from unidimensional numerical scales (e.g. Likert) and categorical judgments to sophisticated statistical and model-based semantic assessments \citep{Landis1977-fa,Zhang2019-iz,Likert1932-hp}. While numerical scales, such as Likert \citep{Likert1932-hp}, offer apparent simplicity for evaluating LLM-generated text, this very simplicity can lead to an oversimplification of complex response attributes, resulting in a loss of critical nuanced information \citep{Westland2022-wi}. Conversely, categorical scoring systems, while providing more distinct classifications, sacrifice granularity in assessment. Statistical and model-based metrics, while offering objectivity and scalability, may not fully capture the qualitative dimensions of LLM responses, such as creativity or nuanced understanding.  Therefore, a balanced evaluation framework is crucial to accurately assess the diverse capabilities of LLMs.

Currently for health evaluations, there is a strong reliance on extensive human expert review (e.g., Khasentino et al.\citep{Cosentino2024-fs} requiring hundreds of hours of expert time to evaluate 900 case studies). While these human evaluations may provide nuanced assessments, they inherently restrict scalability due to high costs and protracted timelines. Furthermore, the subjective nature of Likert scale ratings introduces substantial inter-rater inconsistencies, compromising result reliability \citep{Elangovan2024-ao}. Existing automated evaluation frameworks, while addressing scalability, struggle to replicate the comprehensive, context-aware judgments of human experts, often producing sub-optimal and incomplete evaluations \citep{Shankar2024-kf}. On one hand, LLMs can be capable evaluators and provide evaluations on complex rubrics that align with human evaluators \citep{Chiang2023-gl}.  However, there is the need for specific verification and optimization in safety-critical domains such as health. The conventional paradigm of evaluating complex biomedical models, particularly those integrating blood biomarkers and wearable data with open-ended textual responses, faces critical limitations. This necessitates the development of a novel framework that bridges the gap between human expertise and computational efficiency.

The various aspects of generative AI evaluation may be organized into a framework or taxonomy to facilitate their use. However, available frameworks have a limited focus on principles for evaluation and/or narrow scopes, such as focusing on what, where, and how to evaluate ~\citep{Chang2023-mf}, or providing disparate pathways for specific use  \citep{Guo2023-bg} such as health conversations \citep{Abbasian2024-bg}. Others have suggested a more holistic evaluation, going beyond model evaluation in isolation, and called for understanding the impacts of generative models on humans, society, the economy, and the environment \citep{Weidinger2023-ij}. Recent work has provided a principle-based framework for large language model evaluation by humans in healthcare, including recommendations to assess information quality, reasoning and understanding, expression style and persona, safety and harm, and trust and confidence \citep{Tam2024-qb}. While such a framework is valuable for assessing LLMs in healthcare, it lacks broader applicability to other domains and does not include support for autorater evaluation which may have similar effectiveness to human evaluation in certain domains \citep{Vu2024-qp}.

In this work we introduce the \emph{Adaptive Precise Boolean} rubrics as an evaluation paradigm for scalable health evaluations. We hypothesize that a small set of granular, boolean (Yes/No) criteria enhances consistency and efficiency in complex query evaluation. Existing work has studied the “granularization” of complex evaluation criteria with open-ended or Likert rubrics into a larger set of more focused, and concrete criteria that feature boolean rubrics \citep{Zhong2022-zm,Min2023-hl,Lee2024-tw,Elangovan2024-ao}. These studies have shown improvements in rater reliability and evaluation scoring for tasks such as evaluating summarization, dialogue, and factuality of natural language in general domains without user personalization. Another direction of research shows that learning binary decision trees which can capture complex decision boundaries leads to more explainable decisions \citep{Dasgupta2020-vx}. In \citet{Dasgupta2020-vx}, the decision boundary is $k$-means cluster assignments and the binary decision tree is based on interpretable input features, whereas in our case the decision boundary is a complex evaluation and the boolean rubrics are akin to the nodes in a binary decision tree. Our approach is primarily adapted from \citet{Min2023-hl, Lee2024-tw}. We provide an extended discussion of our design choices in comparison to these works in the \textit{Discussion} section.

We extend the frameworks presented in these works by studying evaluation in the health domain while also accounting for user personalization with health data in both responses and evaluations. Utilizing metabolic health queries, we demonstrate that our proposed framework (1) substantially reduces inter-rater variability between expert and non-expert evaluators compared to traditional Likert scales, (2) halves evaluation time for both groups, (3) achieves automated evaluation parity with expert human judgment, and (4) is more sensitive to missing personal health data and more accurately detects quality discrepancies in responses when integrated with real-world wearable, biomarker, and contextual data. This approach offers a significant advancement in scaling and streamlining large language model evaluation, particularly in specialized domains like health, by improving both accuracy, relevance and cost-effectiveness. While LLMs hold promise for various health applications, this paper specifically addresses the critical need for robust evaluation methodologies and does not present the models discussed herein as approved medical devices or solutions. Real-world deployment of such technologies would necessitate rigorous testing, validation, and regulatory approval.

\section*{Results}
\label{sec:results}
\phantomsection
To address the challenge of evaluating open-ended responses to health-related queries with user personalization, we employed a framework for rubric validation (presented in Figure \ref{fig:figure1}). Utilizing a diverse dataset encompassing representative user queries (Supplemental Data 1), relevant user data (Supplemental Data 2), and targeted prompts (Supplemental Data 3), we generated responses for evaluation from Gemini \citep{2023-my}, a state-of-the-art family of large language model. We subsequently evaluated these responses using three rubric types: (i) established Likert scales, (ii) simple boolean questions based on the initial Likert scales, and (iii) an intelligently sampled set of boolean questions, thus enabling a comparative analysis of their respective strengths and limitations.

\textbf{Rubric Design Framework.} We first used an iterative process to transform rubric criteria characterized by high-complexity response options (e.g., open-ended text or multi-point Likert scales) into a more granular set of rubric criteria employing simplified, binary response options (i.e., boolean “Yes” or “No”); we refer to this approach as \emph{Precise Boolean} rubrics. The primary objective in developing the Precise Boolean rubrics, in contrast to open-ended or Likert-based evaluation methodologies, was to enhance inter-rater reliability in annotation tasks and to generate a more robust and actionable evaluation signal, thereby facilitating programmatic interpretation and response refinement.

Our investigation operated within a framework where evaluation criteria, represented as standardized rubric questions posed to evaluators, remained constant, thus constraining the variability observed in our experiments to the inherent subjectivity and differences between annotators and automatic evaluation (auto-eval) methods. As such, our framework builds on the idea that granular and specific rubric criteria with fewer answer choices will lead to improved inter-rater reliability and allow for automatable action based on responses. As an example, a score of 4 out of 5 on a Likert scale for a rubric criteria which asks about whether user data is used correctly obfuscates the reasoning behind the score itself and cannot be automatically acted on by a larger system without further human intervention. On the other hand, a score of “Yes” or “No” for a rubric criteria which asks about whether a user’s LDL cholesterol data is used correctly provides a clearer evaluation signal and is a better candidate to be programmatically acted on in an evaluation system.

Due to the granular nature of our rubric design, the resulting Precise Boolean rubrics consisted of a substantially larger number of evaluation criteria compared to the starting Likert-scale rubrics (Figure \ref{fig:figure1}D). While auto-eval techniques are well equipped to handle the increased volume of evaluation criteria, the completion of the proposed Precise Boolean rubrics by human annotators presented a considerable practical challenge, potentially rendering manual annotation prohibitively resource-intensive. To mitigate such potential evaluation burden, we further refined the Precise Boolean approach to dynamically filter the extensive set of rubric questions, retaining only the most pertinent criteria, conditioned on the specific data being evaluated.  This data-driven adaptation, referred to as the Adaptive Precise Boolean rubric, enabled a reduction in the total number of evaluations required for each LLM response. This is because user queries and corresponding LLM outputs often exhibit a focused topicality, thus requiring evaluation against only the subset of rubric criteria relevant to those specific themes. For instance, when evaluating an LLM response to a user query concerning clinical risk of diabetes, incorporating rubric criteria related to ferritin could be unnecessary and omitted. 

\begin{figure*}
    \centering
    \includegraphics[width=1.0\textwidth]{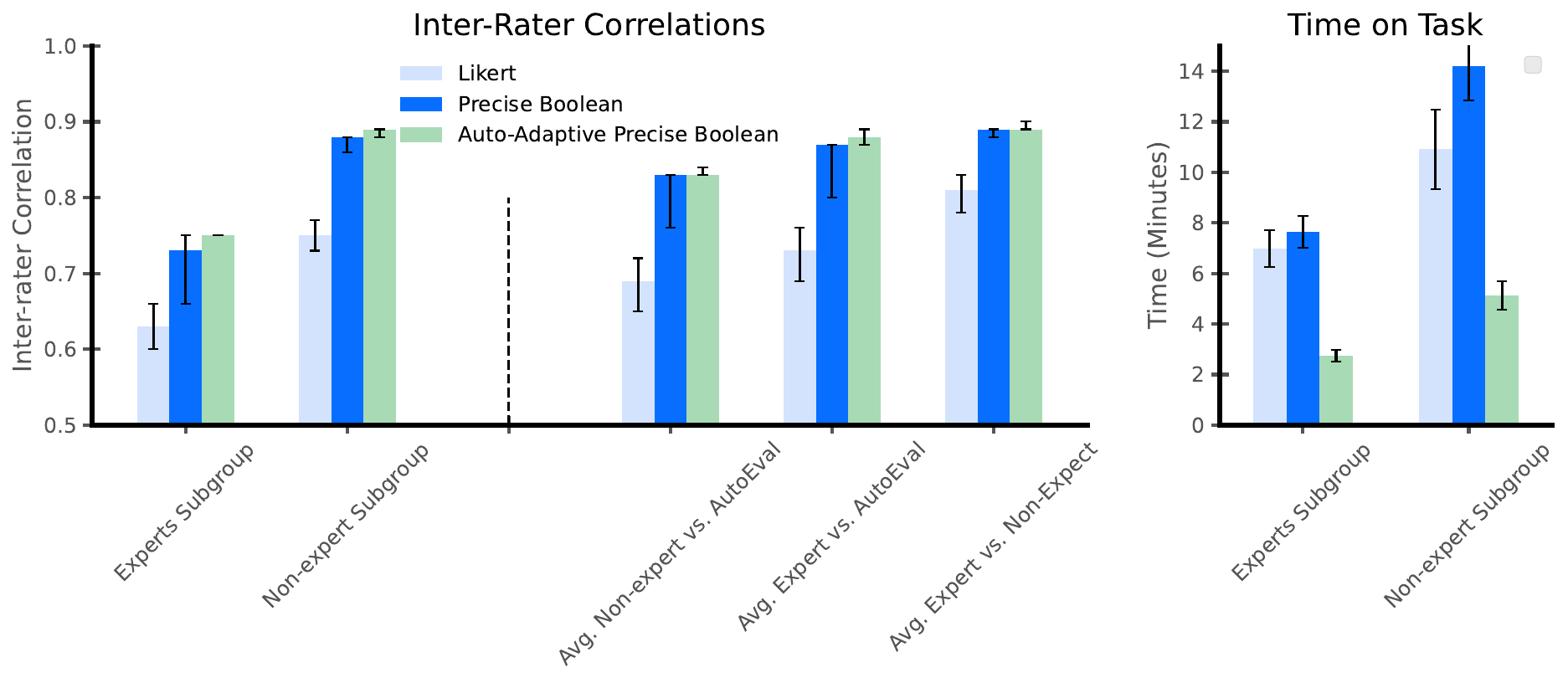}
    \caption{\textbf{Precise Boolean and Adaptive Precise Boolean rubrics increase the consistency between human evaluators (expert and non-expert), and human and automated evaluation.} (A) Inter-rater correlation, as measured by intraclass correlation coefficient (ICC), between different subgroups (human evaluators “expert and non-expert”) and automated evaluation. (B) Adaptive precise rubrics take about half the time needed to do evaluation compared to Likert scale questions.}
    \label{fig:figure2}
\end{figure*}

\begin{figure*}
    \centering
    \includegraphics[width=1.0\textwidth]{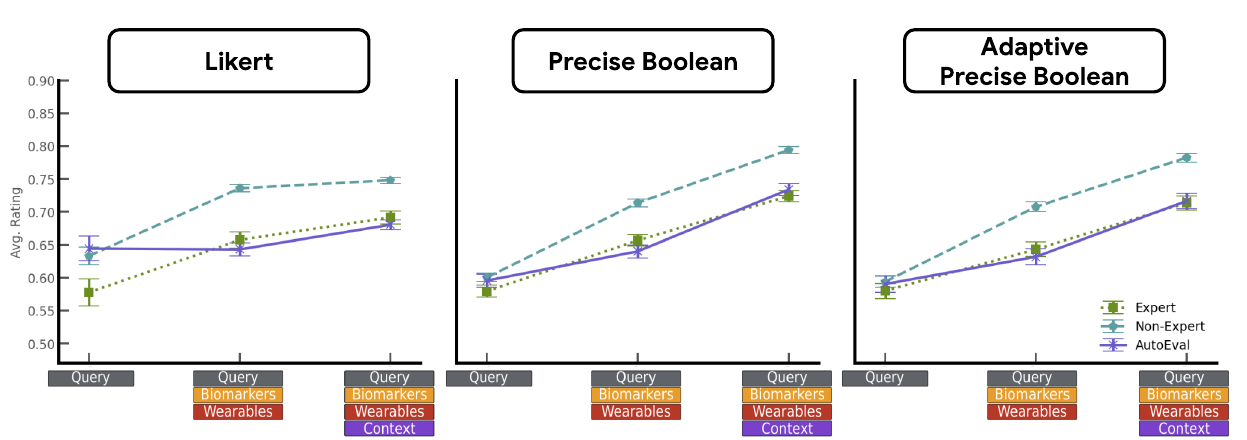}
    \caption{\textbf{Implications on average ratings.}  Ratings obtained from auto-evals using the boolean rubrics are more consistent/correlated with human ratings. In addition, our results show that replacing all questions with an adaptive set has little impact on the evaluation signal.}
    \label{fig:figure3}
\end{figure*}

To convert the Precise Boolean rubrics to Adaptive Precise Boolean ones, we leveraged Gemini as a zero-shot rubric question classifier. That is, the LLM is tasked with determining the relevance of each rubric criterion to a given user query and LLM-generated response pair. Input to the LLM includes the user query, the corresponding LLM response under evaluation, and a specific rubric criterion. The LLM then outputs a binary classification: `1' if the criterion is relevant for evaluating the (query, response) pair, and `0' if not. We refer to rubrics obtained in this manner as Auto-Adaptive Precise Boolean rubrics. To validate this adaptive approach, we established a ground-truth dataset through rubric question classification annotations provided by three medical experts, with majority voting employed to determine the consensus annotation. See \revisiontwo{\methods for} further details on the dataset and consensus metrics for the majority vote. Rubrics obtained based on using this ground-truth dataset in order to do adaptation are referred to as Human-Adaptive Precise Boolean rubrics. As illustrated in Figures \ref{fig:figure2} and \ref{fig:figure3}, the proposed Adaptive Precise Boolean rubrics retain the reliability and scalability advantages of their non-adaptive counterparts. Furthermore, Adaptive Precise Boolean rubrics offer improved evaluation efficiency for human evaluators and, in certain instances, enhance signal detection by filtering out irrelevant or noisy rubric criteria specific to the user query and LLM response being assessed.

For the majority of the results we present comparisons of inter-rater reliability, as measured by intra-class correlation coefficient (ICC), as well as average rubric ratings across all rubrics (\textit{Methods}). We measure ICC as compared to Pearson correlation as we operate in the setting where a fixed set of raters (expert, non-expert, and automatic evaluation) score each target (query-response-rubric question pair), and our goal is to measure the consistency of ratings within each group. We additionally report discrepancy score in \revisiontwo{the \results section}, which measures the relative difference of auto-eval rubric scores under prompt alterations. See \revisiontwo{\methods for} additional discussion on ICC measures and discrepancy scoring. 


\begin{table*}
\centering
\caption{\revisiontwo{\textbf{Accuracy of large language models on certification exam questions.} We report accuracy of various state-of-the-art LLMs on the task of zero-shot multiple-choice question-answering for two benchmark datasets based on practice medical board exam questions in the topics of cardiology and endocrinology. Additionally, we report performance on the difficulty categories (easy, medium, and hard) as determined by the examination source.}}
\resizebox{\textwidth}{!}{%
\revisiontwo{
\begin{tabular}{@{}lllllllll@{}}
\toprule
 & \multicolumn{4}{c}{\textbf{Cardiology (Accuracy - \%)}} & \multicolumn{4}{c}{\textbf{Endocrinology (Accuracy - \%)}} \\ \midrule
 & \multicolumn{1}{c}{\textbf{\begin{tabular}[c]{@{}c@{}}All \\ ($n = 399$)\end{tabular}}} & \multicolumn{1}{c}{\textbf{\begin{tabular}[c]{@{}c@{}}Easy \\ ($n=100$)\end{tabular}}} & \multicolumn{1}{c}{\textbf{\begin{tabular}[c]{@{}c@{}}Mod. \\ ($n=199$)\end{tabular}}} & \multicolumn{1}{c}{\textbf{\begin{tabular}[c]{@{}c@{}}Hard \\ ($n=100$)\end{tabular}}} & \multicolumn{1}{c}{\textbf{\begin{tabular}[c]{@{}c@{}}All \\ ($n=570$)\end{tabular}}} & \multicolumn{1}{c}{\textbf{\begin{tabular}[c]{@{}c@{}}Mod/Easy \\ ($n = 199$)\end{tabular}}} & \multicolumn{1}{c}{\textbf{\begin{tabular}[c]{@{}c@{}}Hard \\ ($n = 198$)\end{tabular}}} & \multicolumn{1}{c}{\textbf{\begin{tabular}[c]{@{}c@{}}Expert \\ ($n = 173$)\end{tabular}}} \\ \midrule
\textbf{Gemini 1.5 Flash} & 59.18 & 75.16 & 59.15 & 42.47 & 69.42 & 74.00 & 68.84 & 66.00 \\
\textbf{Gemini 1.5 Pro} & 69.11 & 86.27 & 67.68 & 52.74 & 75.19 & 75.00 & 72.86 & 80.00 \\
\textbf{Gemini 2.0 Flash} & 70.19 & 85.62 & 71.34 & 52.74 & 77.44 & 79.00 & 76.38 & 78.00 \\
\textbf{Gemini 2.0 Pro} & \textbf{77.54} & \textbf{93.46} & 81.10 & \textbf{56.85} & \textbf{82.21} & 80.00 & \textbf{81.41} & \textbf{86.00} \\
\textbf{GPT-4o} & 75.16 & \textbf{93.46} & 80.49 & 50.00 & 80.95 & 80.00 & \textbf{81.41} & 81.00 \\
\textbf{Claude 3.5 Sonnet} & 74.73 & 87.58 & \textbf{81.71} & 53.42 & 78.70 & \textbf{83.00} & 77.39 & 77.00 \\ \bottomrule
\end{tabular}%
}}
\label{tab:first_table}
\end{table*}

\textbf{Selecting a Large Language Model.} In this work, our focus is to develop a generalizable evaluation framework for LLMs instead of training  a new LLM. In order to select a base model with which to perform the majority of our experiments we tested a set of recent LLMs (Gemini 1.5 Flash/Pro, Gemini 2.0 Flash/Pro, GPT-4o, Claude 3.5 Sonnet). Prior work using GPT and Gemini \citep{Saab2024-bw,Yang2024-do,Cosentino2024-fs} has historically demonstrated strong performance on medical tasks \citep{Kanjee2023-sr}. 

We started our investigation with a closed-ended benchmark task in order to select an LLM family in a more principled manner where evaluation criteria is clear-cut (in this case, accuracy) before moving to our primary setting in which we examine more complicated evaluation criteria. We considered two curated multiple choice question (MCQ) datasets, encompassing 969 questions, to evaluate the performance of language models in answering expert-level health-related questions. These datasets are targeted at evaluating model proficiency in well-studied areas of healthcare, representing domains where expert knowledge is particularly critical. Furthermore, these benchmark tasks provided us an initial signal as to whether these LLMs have sufficient background knowledge in the health domain to be considered useful on the tasks of personalized health LLM response generation and evaluation.

Below we detail the specifics of each dataset:

\begin{itemize}
    \item \textbf{Endocrinology Examination}: Leveraging StatPearls’ \textit{American Board of Internal Medicine: Endocrinology, Diabetes, \& Metabolism Exam} (EDME) preparatory quizzes (“Ace The Endocrinology, Diabetes, \& Metabolism Exam” \citep{StatPearls}), we curated 570 questions from all available levels of difficulty (173 “Expert”, 198 “Difficult”, and 199 “Moderate and Easy” questions). It is important to note that the pass rate for ABIM’s EDME varies per  exam \citep{abim}. In 2023 and 2024 cycles, 82\% and 85\% of clinicians taking the EDME subspecialty exam passed \citep{abim2}.\\
    
    \item \textbf{Cardiology Examination}: We curated a list of 399 Cardiology Board Certification questions using BoardVitals’ ABIM-based preparatory question bank (“Cardiology Board Review Questions [2025]”  \citep{boardvitals}). Similar to the Endocrinology examination, we selected questions from all difficulty levels (100 “Hard”, 199 “Moderate”, and 100 “Easy” questions). According to ABIM statistics \citep{abim2}, in 2023, 86\% of clinicians taking the Cardiovascular Disease subspecialty exam received a passing score.
    
\end{itemize}

As shown in \revisiontwo{Table \ref{tab:first_table}}, Gemini 2.0 Pro achieved the highest overall accuracy among all tested models: 77.54\% on Cardiology MCQs (+2.38\% from the second best model, GPT 4o) and 82.21\% on Endocrinology MCQs (+1.26\% from the second best model, GPT 4o). Based on our results, we chose to leverage the Gemini family of models (Gemini Team Google, \citet{2023-my}) for our experiments; specifically, we chose to leverage the best performing non-experimental Gemini model, i.e. Gemini 1.5 Pro, to allow for the reproducibility of our experiments and results (At the time of writing this manuscript, Gemini 2.0 Pro was in the experimental stage). Using the Gemini family of models also allowed us to ensure that there was no contamination of the training data, though this was not a very significant concern as a majority of our analyses involve novel datasets; however, it was of high priority to ensure a clear train and test split.

\textbf{Comparing Likert and Precise Boolean Rubrics.} Current evaluation methodologies for LLM performance in health domains frequently employ Likert scales for scoring complex, multifaceted questions pertaining to model outputs on health-related data \citep{Kanjee2023-sr,Cosentino2024-fs,McDuff2023-rc}. Our baseline Likert rubric is developed based on the prior work of \citet{Cosentino2024-fs}  and the rubric criteria guidelines can be found in Supplemental Data 4. We compare this baseline to our Precise Boolean rubrics (see Supplemental Data 5 and 6), which we create in a data-driven manner from the Likert rubric questions through augmenting each question with associated user persona data.

We find that such Likert-based rubrics often exhibit a degree of non-specificity and subjectivity, potentially leading to inconsistencies in human evaluation (e.g., Figure \ref{fig:figure1}). In contrast, our proposed Precise Boolean rubric approach demonstrably enhances inter-rater reliability: our results showed  significantly higher intra-class correlation coefficients (ICC) when employing Precise Boolean rubrics compared to traditional Likert rubrics (Figure \ref{fig:figure1}). See \revisiontwo{\methods for} further discussion on ICC. While the enhanced specificity inherent in the boolean questions contributes to this improvement, it is noteworthy that the Precise Boolean rubric achieves this heightened reliability despite comprising a substantially larger number of individual questions – approximately an order of magnitude greater than comparable Likert-based rubrics. This suggests that the increased granularity afforded by Precise Boolean rubrics, even with a greater number of questions, effectively mitigates subjective interpretation and fosters more consistent and reliable evaluations.

Given that some language models \revisiontwo{may prefer} generations from their own family of models \citep{panickssery2024llm}, we provide results from DeepSeek V3 \citep{deepseekv3}, OpenAI GPT 4o \citep{gpt4o} and OpenAI o3 Mini \citep{openai_o3}, as shown in \revisiontwo{Supplemental Figures 1 and 2}. Across our experiments with non-Gemini autoraters, we observed the similar results and patterns as when using Gemini as the autorater, which suggest generalizability of our findings with other LLMs. Additionally, in Supplemental Data 7, we reported ICC results for the Likert rubric compared to Precise Boolean rubrics after binarizing the Likert responses obtained from human annotators and automatic evaluation methods. We performed this comparison in order to ensure that the ICC results we reported were not skewed by the different number of total ratings that each target can attain (5 on the Likert rubric and 2 on the Precise Boolean rubric). We considered binarizing the Likert scores by mapping a score of 5 to 1 on a boolean scale while every other score would be mapped to 0, as well as by mapping scores of 4 or 5 to 1 and everything else mapping to 0. In both cases we observed that the ICC reported on the binarized Likert rubrics were significantly worse than that of both the original Likert rubrics and the Precise Boolean rubrics.

\textbf{Comparing Precise Boolean and Auto-Adaptive Precise Boolean Rubrics.} A key advantage of employing the proposed Adaptive Precise Boolean rubrics lies in their capacity to substantially reduce the evaluation burden without compromising inter-rater agreement. By dynamically filtering rubric questions based on their relevance to the specific user query and LLM response, we observed a considerable reduction in the number of questions presented to evaluators. Notably, this adaptive approach yielded comparable levels of inter-rater agreement to the original Precise Boolean rubrics, indicating that the removal of irrelevant questions did not introduce noise or negatively impact the consistency of evaluations. Moreover, the time required for evaluation tasks was reduced by over 50\% when utilizing Adaptive Precise Boolean rubrics (presented in Figure \ref{fig:figure2}). This efficiency gain resulted in evaluation times that were not only significantly lower than those associated with Precise Boolean rubrics but also surpassed the efficiency of traditional Likert scale-based evaluations. These findings underscore the potential of the Adaptive Precise Boolean rubrics to enhance the scalability and practicality of LLM evaluation, particularly in resource-constrained settings or when dealing with complex, multifaceted datasets.

\begin{figure*}
    \centering
    \includegraphics[width=0.7\textwidth]{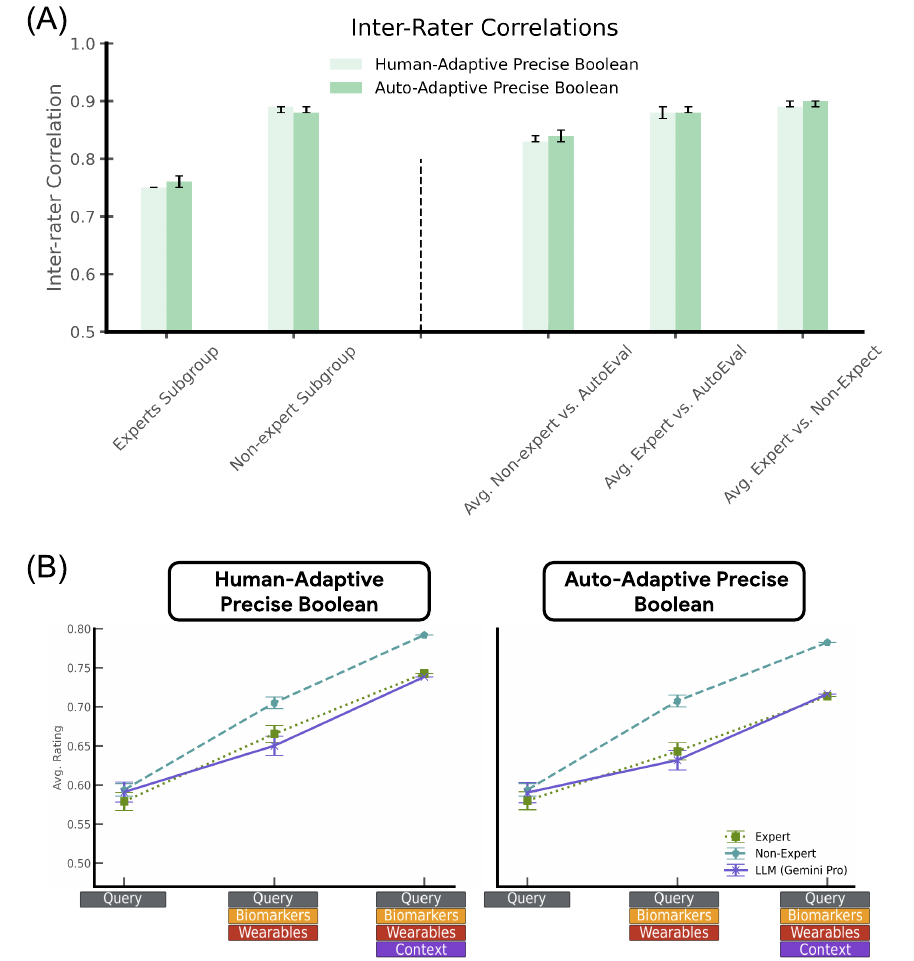}
    \caption{\revisiontwo{\textbf{Comparison of Auto-Adaptive Precise Boolean to Human-Adaptive Precise Boolean rubrics.} (A) Adaptation of Precise Boolean rubrics using Gemini 1.5 Pro as a zero-shot rubric question classifier does not degrade rater correlation metrics (intraclass correlation coefficient, ICC) compared to using human driven adaptation. (B) Auto-Adaptive rubrics show a similar average rating trend to Human-Adaptive rubrics, indicating that the Auto-Adaptive evaluation criteria are sufficient to capture the evaluation signals present based on human adaptation. In the figure legend, ``Gemini Pro" refers to Gemini 1.5 Pro.}}
    \label{fig:figure4}
\end{figure*}

\textbf{The Impact on Average Ratings.} To further evaluate the efficacy of our proposed rubrics, we investigated the sensitivity of both precise and Adaptive Precise Boolean rubrics to variations in response quality. We designed an experiment wherein we systematically augmented user queries with increasing levels of contextual information, hypothesizing that richer queries would elicit higher quality responses from the LLMs under evaluation. As depicted in Figure \ref{fig:figure3}, we observed that average ratings derived from traditional Likert scale rubrics exhibited limited sensitivity to these incremental improvements in input query context. This insensitivity was particularly pronounced in automated evaluations (auto-eval), suggesting a potential ceiling effect or lack of granularity in Likert scales for capturing subtle variations in response quality.

We note that the Likert rubric in our experiments uses a 1 - 5 scale whereas the boolean rubrics report either 0 or 1. To meaningfully compare average rubric scores we considered a normalized average Likert score where the average rubric score computed on the Likert scale was divided by 5 in order to obtain a score in [0, 1].

\begin{table*}
\centering
\caption{\textbf{Results on zero-shot rubric question classification using Gemini 1.5 Pro.} We report accuracy, precision, recall, and F1 metrics on this task comparing model predictions to the ground-truth dataset we constructed over the majority vote of three medical experts on the same task.}
\resizebox{0.6\textwidth}{!}{%
{\begin{tabular}{@{}ccccc@{}}
\toprule
\textbf{Variables} & \textbf{Accuracy \%} & \textbf{Precision \%} & \textbf{Recall \%} & \textbf{F1 \%} \\ \midrule
Total Cholesterol & 75.00 & 81.8 & 75.00 & 78.26 \\
HDL Cholesterol & 75.00 & 76.92 & 83.33 & 80.00 \\
LDL Cholesterol & 75.00 & 81.81 & 75.00 & 78.26 \\
Triglycerides & 75.00 & 75.00 & 81.81 & 78.26 \\
Glucose & 90.00 & 90.91 & 90.91 & 90.91 \\
Hba1c & 75.00 & 100.00 & 58.33 & 73.68 \\
BMI (Body Mass Index) & 85.00 & 81.25 & 100.00 & 89.65 \\
BP (Blood Pressure) & 85.00 & 92.32 & 85.71 & 88.89 \\
Height, Weight, Age & 90.00 & 100.00 & 88.23 & 93.75 \\
Personal Medical History & 95.00 & 100.00 & 94.74 & 97.30 \\
Family Medical History & 90.00 & 100.00 & 88.89 & 94.12 \\
Smoking, Drinking, and Drug History & 60.00 & 60.00 & 81.82 & 69.23 \\
Allergies and Medications & 85.00 & 86.67 & 92.86 & 89.66 \\
Resting Heart Rate, Heart Rate Variability & 50.00 & 62.50 & 71.43 & 66.67 \\
Daily Steps, Total Active Zone Minutes & 75.00 & 84.62 & 78.57 & 81.48 \\
Total Sleep & 55.00 & 69.23 & 64.29 & 66.67 \\
\textbf{All} & \textbf{77.00} & \textbf{83.72} & \textbf{82.57} & \textbf{83.14} \\ \bottomrule
\end{tabular}%
}}
\label{tab:table2}
\end{table*}

\textbf{Comparing Human-Adaptive Precise to Auto-Adaptive Precise Rubrics}. While the Precise Boolean rubrics offer additional granularity and reliability, they inherently comprise a large number of evaluation criteria (due to their binary refinement of each response element). Therefore, for any given user query and LLM response pair, only a subset of these rubric questions are pertinent to the evaluation. This observation led us to investigate whether selecting the most salient components of the Precise Boolean rubrics could be automated without compromising the integrity and discriminative power of the evaluation process. To assess the feasibility of this approach, we provided a preliminary evaluation of a simple classifier used to predict the relevance of individual rubric questions based on the input query and the LLM responses. We then compared the performance of adaptation based on the predictions from this classifier to that of the ground-truth.

To obtain classifications we utilized Gemini as a zero-shot rubric question classifier where user data elements were treated as independent binary classification targets conditioned on the input user query (see \textit{Methods}). As shown in Table \ref{tab:table2}, we found that this method resulted in an averaged accuracy of 77\% on the task of rubric question classification with an averaged F1 score of $\approx$83\%. We observed that the model performed better at classifying relevance of user data such as BMI (body mass index), BP (blood pressure), height, weight, age, and personal and family medical history. On the other hand, the model performed around the level of random guessing on user data such as substance use, total sleep, resting heart rate (RHR), and heart-rate variability (HRV). For blood biomarkers the model was able to obtain an accuracy of $\approx$75\% on all biomarkers we worked with except for blood glucose, which it obtained an accuracy of $\approx$90\% on, suggesting that the model is able to interpret blood biomarker levels to a reasonable degree.

\begin{figure}
    \centering
    \includegraphics[width=0.88\textwidth]{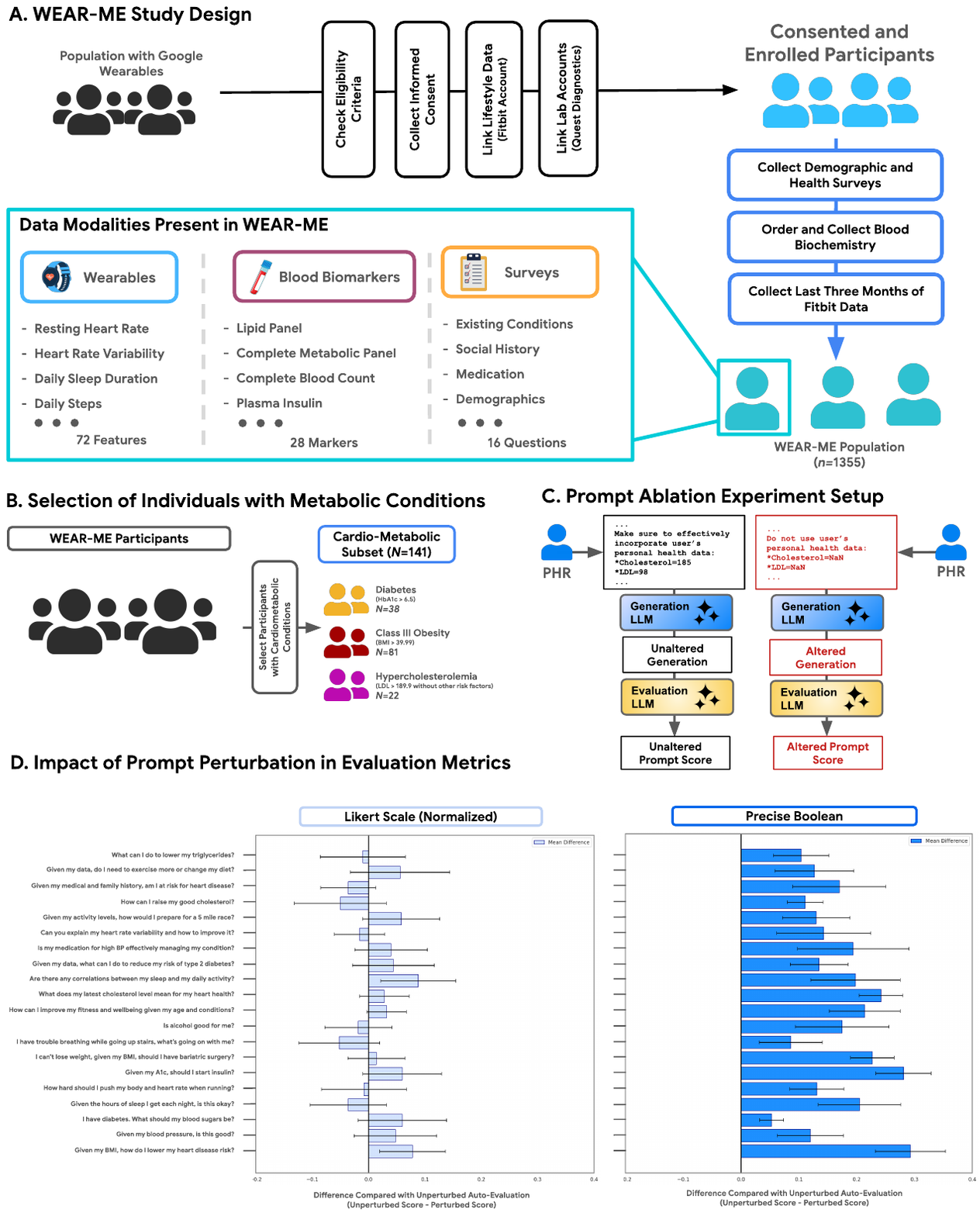}
    \caption{\textbf{Application of proposed approach on a real health study.} (A) Overview of the Wearables for Metabolic Health (WEAR-ME) Study. (B) We filtered participants in the WEAR-ME study based on markers for existing metabolic conditions, particularly obesity (BMI), diabetes (HbA1c), and hypercholesterolemia (LDL) (C) Illustration of our prompt ablation scheme where we altered the generation prompts to not include key blood biomarkers for the incoming queries (D) Measuring the sensitivity of an auto-rater to prompt alterations using Likert rubrics and the proposed Precise Boolean rubrics. Note that the Likert rubrics is normalized (similar to Fig. \ref{fig:figure3}) so that the average discrepancy is on the same scale as the Precise Boolean.}
    \label{fig:figure5}
\end{figure}

In Figure \ref{fig:figure4} we compared ICC metrics and average rubric scores for both Auto-Adaptive Boolean Rubrics and Human-Adaptive Boolean Rubrics. In spite of utilizing an imperfect rubric question classifier, we see that Auto-Adaptive Boolean Rubrics maintain an equivalent improvement of ICC as that of Human-Adaptive Boolean Rubrics. Similarly, the average scores from both rubrics reveals a similar trend as more user data was included in the LLM response generation process. This suggests that the predictions from the rubric question classifier are sufficient to capture the evaluation signal of interest in our setting, which accounts for utilization and correctness of user data in LLM responses as well as indication of harmful or hallucinated response elements. In practice, we would prefer to utilize a rubric question classifier that prioritizes recall over precision so that important user data is not omitted at the risk of over-including some non-relevant rubric questions. We leave it to future work to refine the automatic adaptation methods through fine-tuning or in-context prompting.

\textbf{Application of Precise Boolean Rubrics on Health Data from Real Participants}. To show the applicability and robustness of our proposed approach, we set out to evaluate our framework for detecting abnormalities in the auto-eval stage on deliberately-altered response generations from real data (\textit{Methods}). Specifically, a large-scale (n$\sim$1500) metabolic health study, called Wearables for Metabolic Health (WEAR-ME) \citep{Prieto2024-iu}. As part of the WEAR-ME study, participants linked their Fitbit account (wearables data) to the Google Health Studies (GHS), and asked to grant GHS permission to collect Fitbit data throughout the study, including data for up to 3 months before study enrollment. Once participants were enrolled in the study they were asked to (i) complete four questionnaires which requested demographic information, health history and health information such as sleep and exercise habits, and participant perception of health and previous blood tests (ii) schedule an appointment for a lab test, which was collected by Quest Patient Service Center for a blood draw, and (iii) wear their Fitbit device or Pixel watch during the day and while they sleep (at least 3 out of every 4 days) and track their activities for the duration of the study. Participants were required to review and confirm their blood test results in the GHS App when available. We present an overview of WEAR-ME study design in Figure \ref{fig:figure5}A. WEAR-ME data was handled in accordance with relevant privacy regulations and processes for obtaining informed consent from participants, including the use of their data for research and publication purposes.

To assess the sensitivity of each evaluation framework’s to personalization and contextualization based on the user's health data. For this, we used the same metabolic health queries as before (Supplemental Data 1), which focused on cardiovascular and metabolic health, nutrition, fitness and general wellness. Based on the scope of these questions, we selected participants with confirmed conditions, given the importance of blood and digital biomarkers in responding to these questions for these participants. We selected WEAR-ME participants with metabolic conditions, resulting in 141 participants, 137 of which are unique across all groups (i.e. 4 individuals overlap in more than one group). We present an illustration of the selection criteria in Figure \ref{fig:figure5}B, and describe in detail below. 

\begin{itemize}
    \item[1.] \textbf{Class III (Morbid) Obesity:} As the first group, we select individuals with BMI greater than 40 kg/m$^2$, which constitutes class III (morbid) obesity, resulting in 81 participants. 
    \item[2.] \textbf{Participants with Diabetes:} Per the clinical definition for diabetes, we filtered individuals with HbA1c $> 6.5$, yielding 38 participants.
    \item[3.] \textbf{Hypercholesterolemia (without other major factors):} Using the clinical definition for the most certain case of hypercholesterolemia \citep{cleavland-clinic}, we subsetted individuals with LDL cholesterol $> 190$, resulting in 22 participants.
\end{itemize}

For each participant in the above subset, we then generated responses to the health-related queries with (i) \textit{Unaltered generation prompts:} These prompts were the original prompts used in the above experiments. (ii) \textit{Altered generation prompts:} These prompts were deliberately changed to include NaN values for key biomarkers (clinical biomarkers used to diagnose the underlying condition of the user). The prompt also included explicit generation instructions for the LLM to not leverage the user's personal health data. We present a visual representation of our scheme in Figure \ref{fig:figure5}D.

We evaluated the auto evaluation discrepancy caused by the alterations per each scale (Likert or Precise Boolean) using our defined discrepancy metric (\textit{Methods}). Intuitively, discrepancy between two generations measures how differently an autoeval system rates each response according to the rubrics. In our formulation, a higher positive discrepancy between the unaltered and altered generations is desired. We calculate the discrepancy score (\textit{Methods}) for the Likert and Precise Boolean frameworks, and compute the average discrepancy score across all participants for each query. As shown in Figure \ref{fig:figure5}D, our results indicate that the accuracy score of the Precise Boolean framework is consistently dropped for the perturbed generation compared with the unperturbed generation (since $D_{PB}$ is consistently positive across all queries). In contrast, the average $D_{Likert}$ score across all participants does not follow the same clear pattern (some are positive where a few are negative). Moreover, the absolute magnitude of the $D_{Likert}$ is smaller than $D_{PB}$ across all queries. These results demonstrate that the proposed Precise Boolean framework is more sensitive to inclusion of personal data in the generated responses, which result in more robust auto evaluation pipelines.

\section*{Discussion}
\label{sec:discussion}
\phantomsection
In this study, we proposed a scalable and adaptive evaluation framework specifically designed for health-focused LLMs. Our findings demonstrate that many state-of-the-art LLMs exhibit a notable capacity to: (i) address intricate, medical board examination-style questions within a “closed-book” benchmark environment, (ii) generate personalized and pertinent responses to user health inquiries, effectively incorporating individual health data, (iii) autonomously evaluate and assess personalized health query responses, leveraging rubric criteria as evaluative guidelines; and (4) function as zero-shot rubric question classifiers, capable of dynamically identifying and filtering out rubric criteria irrelevant to specific user queries. Furthermore, we present an iterative approach for transforming traditional Likert-style rubric criteria into a more granular set of Adaptive Precise Boolean rubrics. This transformation yields a substantial improvement in inter-rater reliability across expert clinicians, non-expert evaluators, and automated evaluation methodologies, while also providing a more discerning evaluation signal (in our study the evaluation signal reflects the applicability, correctness, and potential for harm in LLM responses with respect to user personalization in addressing health queries). While the automated components incur computational costs, our findings suggest these are far outweighed by the significant savings in human evaluation time, particularly that of domain experts. Moreover, Adaptive Precise Boolean rubrics maintain the advantages of their Precise Boolean counterparts, concurrently achieving a significant reduction in required evaluation time.

The development of Precise Boolean rubrics in this work stemmed from the fundamental principle of decomposing complex Likert-based evaluation questions, as used in current evaluation approaches, into a more extensive and granular set of binary questions. We operated under the hypothesis that this iterative decomposition process effectively shifts the locus of complexity in evaluation tasks. Rather than burdening evaluators with subjective interpretations and nuanced judgments required by Likert-like rubrics, the complexity is strategically transferred to the upfront design and refinement of the evaluation rubrics themselves. This approach prioritizes the articulation of clear, objective, and easily verifiable criteria, thereby enhancing the consistency and reliability of subsequent evaluations. This shift can be conceptualized as an “evaluation decision-tree”: if subject-matter experts can effectively codify their domain expertise and evaluative reasoning into a binary decision-tree framework, then the transition from Likert to Precise Boolean rubrics can be viewed as leveraging the nodes within such an evaluation decision-tree as the fundamental rubric questions. 

In this framework, boolean responses act as directional signals, guiding the traversal through the decision tree, each binary response dictating the path taken, progressively narrowing down the evaluative space until a definitive and concrete assessment is reached. This structured approach not only enhances the transparency and explainability of the evaluation process but also facilitates the identification of specific areas of strength or weakness in the LLM responses, moving beyond a singular, often ambiguous, Likert score. While our framework does not eliminate rater subjectivity, it fundamentally constrains it. Unlike a Likert scale, where ambiguity exists in both the interpretation of the criteria and the meaning of the scale points (e.g., the difference between a `4' and a `5'), our approach isolates subjectivity to a single binary decision on a highly specific criterion. We posit that this reduction in the `surface area' for subjective interpretation is a key driver of the substantially higher inter-rater reliability observed in our experiments (Fig. \ref{fig:figure2}).

Beyond the conceptual refinement of evaluation rubrics, our work also explored data-driven methodologies to streamline the curation of the Precise Boolean rubrics, effectively bootstrapping their development even prior to the involvement of domain experts. This idea was based on the observation that Likert-like questions could be readily transformed into a more extensive set of Precise Boolean rubrics by incorporating the specifics of user data available during the evaluation stage. This data-driven expansion leverages the inherent structure of personalized health information to automatically generate more comprehensive and targeted rubric sets, significantly reducing the initial manual effort typically required in rubric design. Subsequent validation revealed that rubrics generated through this method enhanced inter-rater reliability—demonstrated by significantly improved score correlations within and between expert, non-expert, and automated evaluation subgroups— as well as yielding more accurate reflection of the underlying evaluation signals. This suggests that the data-driven generation of Precise Boolean rubrics not only improves the consistency of evaluations across diverse rater groups but also serves to clarify and emphasize the core evaluation criteria, leading to a more robust and insightful assessment of LLM performance in personalized health applications.

In comparison to prior work, Min et al.\citep{Min2023-hl} and Lee et al.\citep{Lee2024-tw} utilize LLMs to expand the initial set of evaluation criteria into a larger set of boolean criteria. On the other hand we employ a template-driven method to create our initial boolean criteria based on user personalization data elements in our dataset(s). \citet{Min2023-hl} further iterate their process using human annotators, which we also do through medical expert feedback. Both prior works then employ human annotation to determine whether newly created, granularized evaluation criteria are relevant or irrelevant to a given prompt or topic. In contrast we utilize Gemini as a zero-shot rubric question classifier in order to automatically filter the final set of Precise Boolean rubrics to create our Auto-Adaptive Precise Boolean rubrics. We note that concurrent updates to the initial manuscript of \citet{Lee2024-tw} have now incorporated LLM-driven methods for boolean question filtering. Further, \citet{Lee2024-tw} utilize general domain datasets for evaluation that focus on news article summarization \citep{Fabbri2020-rx} and dialogue about topics such as fashion, politics, books, and sports, to name a few \citep{Gopalakrishnan2023-qi}, whereas \citet{Min2023-hl} creates a dataset through summarizations of Wikipedia pages on people. Both settings differ from ours in that we study a more focused and specialized domain of health LLM responses to user queries, and our data requires accounting for user personalization with multi-modal user data (blood biomarker readings, time-series wearable data, and user health context) in the generation process and in the response evaluation process. We see our results as a necessary and complementary direction of study in relation to prior work on creating “atomic” or granular rubric criteria for more efficient and reliable evaluation processes.

To validate the practical applicability and robustness of our proposed Precise Boolean auto-evaluation framework, we conducted a large-scale experiment using data from the WEAR-ME metabolic health study. This experiment was designed not only to test the sensitivity of different auto-evaluation rubrics but also to simulate a common real-world failure mode: the presence of incomplete or ignored personalized health data. Critically, we introduced deliberate alterations by removing key biomarkers from the generation inputs, simulating a scenario where the LLM either has data gaps or fails to utilize a user's health records. We then quantified the discrepancy in auto-evaluation scores between responses generated with unaltered (with all data) and altered (removed data) prompts.

Our findings demonstrate a clear and consistent advantage for the Precise Boolean framework. As shown in Fig. \ref{fig:figure5}(D), the Precise Boolean rubric consistently detected a significant drop in response quality when personal data was withheld (indicated by the large, positive discrepancy scores across all queries), highlighting its heightened sensitivity to the correct use of contextual information. This finding is particularly relevant in health contexts where data quality can vary significantly. While a traditional Likert rubric might also yield a lower score, its response to the same data perturbations was less consistent and significantly weaker. The ambiguity of a single Likert score makes it difficult to diagnose the source of the quality degradation. In contrast, our framework provides a more actionable, diagnostic signal by effectively pinpointing failures in data utilization. This capability is essential for building robust and reliable evaluation pipelines, particularly for sensitive domains like healthcare where the accurate use of personal data is paramount for generating safe and effective responses.

It is important to acknowledge and seek to address potential biases and the ethical dimensions of the proposed evaluation framework. While our results demonstrate a significant reduction in inter-rater variability compared to traditional Likert scales, this does not eliminate the risk of annotator bias, which can be influenced by factors such as cultural background, expertise, and implicit assumptions \citep{clark-2021}. For instance, previous work on health equity explicitly highlights how biases can manifest as inaccuracies across identity axes, stereotypical language, or the omission of structural explanations for health inequities \citep{pfohl-2024}. Within this context, we posit that while rater bias may manifest similarly across both Likert and Boolean scales, our framework offers a significant advantage in its detection and localization. A subjective judgment on a 5-point Likert scale is inherently ambiguous, making it difficult to disentangle the source of disagreement or bias. In contrast, by decomposing a complex evaluation into a series of precise, atomic Boolean questions, our framework can pinpoint the exact criteria where rater subjectivity or bias is most pronounced. This granularity transforms the abstract problem of bias into a more tractable issue by localizing it to specific, verifiable aspects of a model's response.

Furthermore, the high inter-rater reliability achieved with Boolean rubrics provides a strong statistical foundation for identifying biased evaluations. High agreement establishes a clear consensus, allowing for the detection of outlier raters whose response patterns deviate significantly from the group. Consequently, inter-rater reliability can serve as a valuable proxy for the ability to de-noise biased ratings from a larger pool of evaluators. By averaging ratings across a diverse rater population, the influence of individual biases can be attenuated, yielding an aggregate score that, in expectation, is a more robust and unbiased estimate of response quality. Nonetheless, the simplification of complex medical information into Boolean rubrics, while efficient, risks oversimplifying nuanced clinical advice, potentially leading to the misinterpretation or omission of critical context. This is a recognized challenge in health AI, where the failure to capture semantic meaning can have significant consequences \citep{Abbasian2024-bg}. Future iterations of this framework should incorporate methods for actively identifying and mitigating these biases, for instance through targeted rater training on health equity principles \citep{gehrmann-2022}, and by exploring hybrid evaluation models that combine the scalability of Boolean rubrics with qualitative assessments for particularly ambiguous cases.

A primary direction for future work is to enhance the framework's capacity to handle the inherent ambiguity and context-dependency of complex health responses. While the binary format of Precise Boolean rubrics is designed to maximize clarity and reliability, we acknowledge that some nuances may be challenging to capture in a strictly ``Yes''/``No'' format. Our framework is designed to address this not as a static limitation, but through an iterative, human-in-the-loop refinement process. Augmenting the Boolean rubric with an open-ended comment field could serve as a simple mitigation strategy: When an evaluator encounters an ambiguous or partially correct response that defies a simple binary judgment, they can provide qualitative feedback, which could serve as a critical input for the next cycle of rubric development. A human or automated expert system would then interpret this feedback to identify the source of ambiguity and propose an expanded or refined set of more atomic Boolean criteria that explicitly account for the identified edge case or context. For example, a single ambiguous question could be decomposed into several context-dependent sub-questions that are answerable with clear Boolean logic. This iterative cycle, a core principle of our proposed methodology, transforms ambiguity from a persistent problem into an opportunity to progressively enhance the precision and coverage of the evaluation rubric. This adaptability ensures the framework remains relevant over time, providing a stable yet evolving standard against which new model versions can be longitudinally assessed for both improvements and regressions.

A significant avenue for future research involves exploring how this evaluation framework can be integrated into model training and fine-tuning pipelines, such as Reinforcement Learning from Human Feedback (RLHF). The granular, interpretable signal provided by Precise Boolean rubrics could serve as a high-quality reward for optimizing model behavior. We hypothesize that fine-tuning against these criteria could effectively steer models towards greater factuality, safety, and adherence to verifiable instructions. However, a critical research challenge will be to balance this optimization against more holistic, qualitative aspects of a response, such as conversational tone and empathy. Investigating this trade-off and preventing models from simply "teaching to the test" by satisfying a checklist at the expense of overall quality is a crucial next step for responsibly leveraging fine-grained evaluation in model development.

Furthermore, we aim to investigate the robustness of the auto-evaluation component. This includes exploring the framework's resilience against adversarial manipulations, where an LLM being evaluated might generate responses specifically designed to exploit the auto-rater's logic \citep{yu2025}. Developing methods to detect and mitigate such behaviors will be crucial for deploying this framework in high-stakes settings. Another potential future direction is to improve the rubric question classification process and explore alternative data-driven or model-driven methods for breaking apart Likert evaluation questions. In our work we used user data and subject-matter experts to iterate our framework, however as shown in Lee et al. \citep{Lee2024-tw}, we expect that incorporating LLMs and reasoning models will improve the automatic scalability of our framework in the initial phase of splitting up rubric criteria into a series of Precise Boolean rubrics.

Lastly, we note that this research focuses on the evaluation framework itself and does not present the health language models discussed as approved or ready-for-clinical-use medical devices or solutions. Real-world deployment of such technologies would necessitate rigorous testing, validation, and regulatory approval.
\section*{Methods}
\label{sec:methods}
\phantomsection

\textbf{Evaluation Framework – Precise Boolean Rubrics.} Our starting point for Precise Boolean rubrics is to take an existing open-ended or Likert rubric question. In this work we build off the prior defined rubrics in PH-LLM \citep{Cosentino2024-fs} which utilize a Likert scale. For each rubric question we automatically create many variations on the same question which can be answered using boolean (yes or no) responses. In the case of health LLMs which operate on personalized user data we start by breaking out each user data element into a separate question which can be more precisely addressed by the evaluators.

For example, the rubric question for “This section references all important user data needed.” would be broken down, automatically, into a series of questions that depend on the input user data that the model is provided in the first place. If cholesterol information is tracked for users then a more precise rubric question would be “This section references all important user data needed regarding cholesterol levels”. Finally, rather than asking for evaluation responses on a scale of 1 - 5 we simply ask evaluators to respond with “Yes” or “No” as a boolean scale.

Iterating this process allows us to break down a simple question that involves more complex, open-ended or Likert scale responses into a set of more complex questions that involve a simpler, boolean response. In the construction of our Precise Boolean rubric, we further sought medical expert feedback on the rubric questions in order to break them down further into more precise questions. 

To do this we provided experts with a few random example (query, response) pairs coupled with an initially proposed rubric and asked them to begin filling out the rubric. In any case where the Yes/No response was insufficient for the expert given a rubric criteria, they were asked to propose an alternative set of Yes/No questions that could replace the initial rubric criteria and allow them to provide the level of granular feedback they desired.

\textbf{Evaluation Framework – Adaptive Precise Boolean Rubrics.} As we have seen, the method for our evaluation framework takes one open-ended or Likert scale rubric criteria and, through automatic (data-driven) and human feedback, breaks out that criteria into a set of numerous rubric criteria which can all be answered through simpler, boolean responses. In doing so, however, the newly proposed Precise Boolean rubrics may end up containing significantly more rubric criteria than before. While this may be feasible when employing automatic evaluation methods, and indeed we see in our results that the shift from Likert to Precise Boolean improves the reliability of automatic evaluation methods, it can be prohibitively expensive when human evaluators are needed to answer many more evaluation questions.

To address this we further propose Adaptive Precise Boolean rubrics, in which the full set of Precise Boolean rubric criteria is filtered in an adaptive manner conditioned on the input user query (and optionally the input user data or any other information in the LLM prompt). We find that after iterating on the process of creating a Precise Boolean rubric from a Likert rubric, many of the newly created rubric criteria will be highly specific and pertain only to a subset of user queries that relate to those specific topics. Therefore we can adapt the total set of boolean criteria to the query and only require feedback on those rubric elements deemed relevant.

In our work we explored both human-driven methods and automatic methods of adaptation. In the human-driven method we obtained ground-truth labels for each user query and Precise Boolean rubric element by three medical experts. A majority vote was then taken to create the final filtering matrix. In the automatic evaluation method we experiment with a state-of-the-art LLM (Gemini 1.5 Pro) as a rubric question classifier to create the final filtering matrix in a zero-shot manner. Further details of the rubric question classification process using automatic evaluation methods is provided in the Automatic Evaluation Setup section of Methods.

\textbf{Human Evaluation Setup (Medical Expert vs Non-Expert).} To conduct human evaluations to compare our Likert rubrics with Precise Boolean and Adaptive Precise Boolean rubrics, we first selected a random sample of 20 user queries pertaining to topics such as metabolic health, general fitness and activity, and general health (see Supplemental Data 1). For each query we obtained model responses from Gemini 1.5 Pro while prompting the model with varying amounts of user persona information. This included one-shot blood biomarker readings, time-series wearable fitness data, and additional user health context such as family medical history, personal medical history, and substance use, to name a few. 

We gathered a total of four model responses for each user query where each model response was obtained after prompting with differing user data. The first response was based on only the user query and no personalization data, the second response was prompted with blood biomarker readings along with the user query, the third response was additionally prompted with time-series wearable data, and finally the fourth response also included user context such as medical history, family medical history, and more (see Supplemental Data 3 for a complete description of the synthetic user persona data used in prompting). 

This process resulted in a total of 80 (query, response) pairs. For each (query, response) pair we obtained human rubric evaluations for two rubrics: Likert and Precise Boolean. Adaptive Precise Boolean rubric scores were obtained automatically by subsetting responses from the Precise Boolean rubric according to the rubric question classification results (or the ground-truth majority vote results from \revisiontwo{three medical experts}).

In both expert and non-expert evaluator subgroups we performed two rounds of evaluation, each on a disjoint subset of 10 user queries. We ordered the evaluation tasks on the first subset of 10 queries as follows:
\begin{itemize}
    \item[1.] Precise Boolean rubric for queries 1 - 5,
    \item[2.] Likert rubric for queries 6 - 10,
    \item[3.] Likert rubric for queries 1 - 5,
    \item[4.] Precise Boolean rubric for queries 6 - 10.
\end{itemize}
The same ordering scheme was done for queries 11 - 20 in the second round of annotations. 

This ordering was done to limit biasing effects in the study where evaluators might be influenced by having completed one rubric completely on all queries first before working on the next. In this way, the Likert rubric was completed before the Precise Boolean rubric on half of the queries, and vice versa for the other half of the queries. We additionally randomized the ordering of all (query, response) pairs in each subset before annotation to avoid biasing results due to response quality changing in a specific way as responses were generated with increasing amounts of user data.

All annotations were completed over the course of three months due to expert and non-expert availability. The Likert rubric sheets involved filling in a score on a scale of 1 - 5 for each rubric criteria, while the Boolean rubric sheets involved checking boxes for rubric criteria when they are satisfied by the response. Medical experts and non-expert annotators were provided the same rubric criteria guidelines for both Likert and Precise Boolean rubrics as well as the current accepted standard ranges for blood biomarker readings (\url{https://emedicine.medscape.com/article/2172316-overview.}) and the complete user persona details used when prompting the model. Annotations were completed by 5 medical experts and 10 non-expert annotators.

\textbf{Automatic Evaluation Setup – Auto-Eval of Rubric Criteria.} Automatic evaluation for rubrics was performed using Gemini 1.5 Pro. The model was prompted with the same information as the human experts and non-experts were provided (rubric criteria guidelines, standard ranges for blood biomarker readings, and complete user persona details). The model was further instructed to perform the task of evaluation and scores were obtained on a single (query, response, rubric criteria) tuple at a time.

We experimented with prompting the model to only respond with the numerical rubric score for a given rubric criteria, as well as responding with the numerical score and an explanation of why the score was chosen. We further experimented with two prompt introductions: one that prefaced the model as an “expert medical evaluator”, and another that did not. For both rubrics, the model showed high self-consistency (see Supplemental Data 8) in these evaluations across all four different prompt variations for automatic evaluation. Given this, we opted to perform the remaining automatic evaluations using a single prompt choice which prefaced the model as an “expert medical evaluator” and asked the model to output only a numerical score for the rubric criteria without explanation.

The final automatic evaluation prompts used are available in Supplemental Data 9.

\textbf{Auto-Eval as a Rubric Question Classifier.} For rubric question classification we utilized Gemini Pro 1.5 in a zero-shot fashion. To do this we prompted the model with a user query and a user data or user persona element used to augment rubric criteria in the Precise Boolean rubrics. We then asked the model to output a 1 if the user data would be relevant to create a response or to evaluate a response for a given user query, and a 0 if not. Given that we constructed Precise Boolean rubrics in a templated fashion by filling in user data and contextual user persona elements into variations on the Likert rubric base question, we treated rubric question classification as a task of predicting which user data or user persona elements were relevant to a given user query. This task is agnostic to the specific rubric criteria being considered, i.e. “is the data relevant” or “are there hallucinations” as separate rubric criteria would both make use of the same user data in a Precise Boolean or Adaptive Precise Boolean rubric if the data itself was relevant to the user’s base query. 

This scheme enabled us to treat the rubric question classification task as a series of boolean classification problems for each user data element. Accuracy, precision, recall, and F1 metrics were computed over the subset of 20 user queries used in our study and the ground truth was obtained from a majority vote over \revisiontwo{three} medical experts performing the same task. The ground truth majority vote labels for this dataset can be found in Supplemental Data 10.

We further compute the agreement over the three medical experts on each of the rubric question classification tasks over all twenty queries in order to ascertain how robust the labeling process was for this dataset. The agreement scores can be found in Supplemental Data 11. We found that for the blood biomarkers all three medical experts agreed on the ground truth labeling for 16/20 questions, whereas for all of the additional user health questions the three medical experts agreed on all 20/20 questions. Finally, for the time-series wearable data the experts agreed on 19/20 questions. Overall this suggests that in spite of only having three medical experts to take a majority vote over, the resulting labels are highly consistent with each other.

\textbf{Prompt Ablation Experiment.} For each participant in the selected subset, and for each health-related query, we generated two types of prompts:
\begin{itemize}
    \item \emph{Unaltered Prompts ($U$)}:  These prompts contained the health-related query, along with relevant participant-specific data, including demographic, lab tests and wearables-derived features (i.e. “PHR + Activity+Context” prompts). \\
    \item \emph{Altered Prompts ($\Tilde{U}$)}: These prompts were identical to the unaltered prompts except that key biomarker values related to the participant's primary condition were replaced with "NaN" (Not a Number).  Additionally, an explicit instruction was added to the prompt, directing the LLM not to utilize the participant's personal health data.
\end{itemize}

\paragraph{Response Generation} Both the unaltered ($U$) and altered ($\Tilde{U}$) prompts were used to generate responses using Gemini 1.5 Flash. Responses were generated using a temperature setting of 0.6 and a top-p value of 0.95 to ensure diverse and consistent generations.

\paragraph{Measure of Discrepancy.} We define the response discrepancy between unaltered and altered generation prompts as shown in \eqref{eq:discrepency}:
\begin{align}
    D_{\{PB, Likert\}}(Q, R, LLM, U, \Tilde{U}) &= \frac{1}{K} \sum_{i=1}^K AE(Q, R_i, LLM(U)) - AE(Q, R_i, LLM(\Tilde{U}))
    \label{eq:discrepency}
\end{align}
where $Q$ denotes user’s query, $R_j$ denotes the $j^{th}$ rubric ($K$-many rubrics in total), $AE(\cdot)$ refers to an inference call to the auto evaluator, and $LLM(\cdot)$ represents the response generated by the generation LLM with $U$, $\Tilde{U}$ denoting the unaltered and altered prompts, respectively.

\textbf{Prompting LLMs for Multiple Choice Questions.} We prompted all LLMs using a minimal prompt in order to reduce any performance changes based on the prompting technique. Specifically, for questions with four possible choices, we prompted each model using the following dynamic prompts (where the variable inside `{}` was replaced with the appropriate value at execution time):

\begin{tcolorbox}[colback=white, colframe=black, arc=3mm]
\begin{lstlisting}[basicstyle=\ttfamily]
I will ask you a multiple choice question and provide four answer
options lettered 'A', 'B', 'C', or 'D'. Please respond with the 
correct letter first and then a brief explanation of your reasoning 
for the answer.

    Question: {question}
    {a}
    {b}
    {c}
    {d}

    Correct:
\end{lstlisting}
\end{tcolorbox}

 and for questions with five possible answers, we use the prompt below:

\begin{tcolorbox}[colback=white, colframe=black, arc=3mm]
\begin{lstlisting}[basicstyle=\ttfamily]
I will ask you a multiple choice question and provide four answer
options lettered 'A', 'B', 'C', or 'D', or 'E'. Please respond with 
the correct letter first and then a brief explanation of your 
reasoning for the answer.

    Question: {question}
    {a}
    {b}
    {c}
    {d}
    {e}

    Correct:
\end{lstlisting}
\end{tcolorbox}
 
To compute the accuracy of the models, we parsed the output of the LLMs for each question to extract the letter that each model responded with per question. We then compared the extracted letters with the ground truth correct choice, and computed the percentage of correct responses across each dataset. Note that while we could leverage an LLM to evaluate model responses, we chose to do so with our predefined parsing function to reduce model bias (e.g. an OpenAI auto-rater preferring responses from OpenAI LLMs) or mistakes by an auto-eval system.

\textbf{LLM Models Benchmarked}. As part of this work, we benchmarked the following LLMs:

\begin{itemize}
    \item \textbf{Gemini 1.5 Pro:} Gemini 1.5 Pro is a multimodal LLM from Google, notable for its large 1 million token context window. \item \textbf{Gemini 1.5 Flash:} Developed by Google, Gemini 1.5 Flash is a lighter and faster model compared with Gemini 1.5 Pro. It inherits the 1 million token context window from Gemini 1.5 Pro, allowing it to process large amounts of information.
    \item \textbf{Gemini 2.0 Flash and Pro:} These models are the next-generation of Gemini 1.5 models with the same 1 million input context window.
    \item \textbf{GPT-4o:} A multimodal model developed by OpenAI that integrates text, audio, and vision capabilities natively. This model is designed to provide a more seamless and interactive experience across different modalities, suitable for a wide range of multi-modal applications.
    \item \textbf{Claude 3.5 Sonnet:} Developed by Anthropic, Claude 3.5 Sonnet is a multimodal LLM designed for applications requiring multi-modal reasoning and understanding.
\end{itemize}

\textbf{Evaluation Metrics.} For multiple choice question-answering we report standard multi-class accuracy, stratified by question difficulty. To compare human evaluations and automatic evaluations through rubric scores, we calculated intraclass correlation coefficients (ICC) \citep{Fisher1925-zc} within various evaluator subgroups as well as between evaluator subgroups. Indeed, ICC measures are commonly used in prior literature and have been extensively studied to assess the reliability of evaluators, whether human or through automatic evaluation methods \citep{Bartko1966-fe,Shrout1979-yy,Liljequist2019-kh,Hackl2023-zb}. Based on our human evaluation study design, we report results for the “ICC3” metric which is used when the same set of raters annotate each target. ICC scores range from 0 (completely unreliable) to 1 (completely reliable). We computed ICC scores using the \texttt{pingouin} Python package (\url{https://pingouin-stats.org/build/html/generated/pingouin.intraclass_corr.html}). In addition to ICC we also report the average rubric score for (query, response) pairs in order to measure the response quality for a user query conditioned on the rubric.

\textbf{Processing WEAR-ME Data for Prompt Ablation Experiment.} The WEAR-ME dataset \citep{metwally2025-ir} is a dataset comprising wearable sensor data and blood biomarker measurements collected from participants enrolled in a health study. Participants, recruited via a Google Health Studies, consented to share data from their Fitbit or Pixel Watch devices, which included heart rate metrics (heart rate, resting heart rate, heart rate variability), physical activity (steps, active zone minutes), sleep patterns, respiration rate, skin temperature, SpO2, and weight. Concurrently, participants underwent blood draws at Quest Diagnostics, providing a panel of biomarkers including Complete Blood Count, Comprehensive Metabolic Panel, lipid profiles, HbA1c, and inflammatory markers. 

To process the data prior to selecting individuals with metabolic conditions, digital markers derived from Fitbit algorithms were aggregated using mean, standard deviation, and median values over 90 days prior to blood collection. As a quality control measure, participants with implausible BMI values (outside the 12-65 range) or self-reported non-fasting status at blood draw were excluded. Furthermore, for each analysis, individuals with any missing values within the chosen feature set were removed to ensure data integrity and analytical rigor. This data preprocessing pipeline facilitated the integration of multi-modal health data to select for individuals with metabolic health conditions.

From the processed data, we selected participants with metabolic conditions, particularly those with: 
\begin{itemize}
    \item Class III (Morbid) Obesity, comprising 81 participants with a Body Mass Index exceeding 40  kg/m$^2$
    \item Diabetes, encompassing 38 participants meeting the clinical threshold for HbA1c levels (> 6.5\%)
    \item Hypercholesterolemia (absent other major confounding factors), consisting of 22 participants with LDL cholesterol levels surpassing 190 mg/dL. These subgroups, defined by established clinical benchmarks, provided the foundation for targeted analyses of LLM performance in the context of metabolic health.
\end{itemize}
This filtering resulted in 141, including 137 unique individuals across all defined subgroups.

\section*{Data Availability}
All data related to prompts, queries, synthetic personas and evaluation rubrics are provided in the supplementary material. Data access to the WEAR-ME study can be found in Metwally et al. \cite{metwally2025-ir}

\section*{Competing Interests}

 A.A.H., X.L., A.Z.F., B.W., N.H., C.S., M.M, S.P., J.L.P, D.M., and A.A.M. are or were employees of Alphabet at the time of submission, and may own stock as part of the standard compensation package. N.M. was an intern at Google during this research. All other authors declare no competing interests.

\section*{Author Contributions}

N.M., A.A.H., D.M., A.A.M. conceptualized and designed the research. N.M., A.A.H., and B.G. conducted data curation. N.M., A.A.H., D.M., and A.A.M. analyzed and visualized data. N.M., A.A.H., X.L., D.M., and A.A.M. wrote the original draft of the paper. N.M., A.A.H., X.L., A.Z.F., B.W., N.H., B.G., C.S., M.M., S.P., J.L.P., D.M., and A.A.M. reviewed and edited the manuscript. A.A.M. contributed to project administration. D.M., A.A.M. contributed to project supervision.

\section*{Acknowledgement}

This study was funded by Google LLC. We are deeply grateful to the members of the Human Research Laboratory at Google for helping set up evaluation workflows for human evaluators, in particular, Erik Schenck, and Derek Peyton. We thank our expert evaluators Michelle Jonelis, Narayan Krishnamurthy, Thuan Dang, Timothy Wong, and Andreas Michaelides and non-expert evaluators Aayush Ranjan, Pawan, Shwetank Dhruva, and Nitesh Tiwari.

\pagebreak
\bibliography{main_arxiv}

@ARTICLE{Lee2024-tw,
  title    = "{CheckEval}: A Reliable {LLM-as-a-Judge} Framework for Evaluating Text Generation Using Checklists",
  author   = "Lee, Yukyung and Kim, Joonghoon and Kim, Jaehee and Cho, Hyowon
              and Kang, Jaewook and Kang, Pilsung and Kim, Najoung",
  journal = {HEAL Workshop at CHI},
  month    =  mar,
  year     =  2024,
  eprint   = "2403.18771"
}

@ARTICLE{Shankar2024-kf,
  title    = "Who Validates the Validators? Aligning {LLM}-Assisted Evaluation of {LLM} Outputs With Human Preferences",
  author   = "Shankar, Shreya and Zamfirescu-Pereira, J D and Hartmann,
              Bj{\"o}rn and Parameswaran, Aditya G and Arawjo, Ian",
  journal = "UIST '24: Proceedings of the 37th Annual ACM Symposium on User Interface Software and Technology",
  month    =  apr,
  year     =  2024,
  eprint   = "2404.12272"
}

@ARTICLE{Elangovan2024-ao,
  title    = "{ConSiDERS-The-Human} Evaluation Framework: Rethinking Human Evaluation for Generative Large Language Models",
  author   = "Elangovan, Aparna and Liu, Ling and Xu, Lei and Bodapati, Sravan
              and Roth, Dan",
  journal = "Proceedings of the 62nd Annual Meeting of the Association for Computational Linguistics",
  month    =  may,
  year     =  2024
}

@MISC{boardvitals,
  title        = "Cardiology Board Review Questions [2025] - {BoardVitals}",
  author = "{BoardVitals}",
  howpublished = "\url{https://www.boardvitals.com/cardiology-board-review/?utm_term=&utm_campaign=Performance+Max+-+BoardReview&utm_source=google&utm_medium=cpc&hsa_acc=3629361371&hsa_cam=16996727962&hsa_grp=&hsa_ad=&hsa_src=x&hsa_tgt=&hsa_kw=&hsa_mt=&hsa_net=adwords&hsa_ver=3&utm_content=april-flash&gad_source=1&gclid=EAIaIQobChMI7tewscPJiwMVAQCtBh3AiAH2EAAYASAAEgI_S_D_BwE}",
  note         = "Accessed: 2025-3-19",
  year="2025",
}

@ARTICLE{2023-my,
  title = "{Gemini}: A Family of Highly Capable Multimodal Models",
  author   = "{Gemini Team, Google}",
  journal = "arXiv preprint",
  month    =  "December",
  year     =  "2023",
  eprint={2312.11805},
  archivePrefix={arXiv},
  primaryClass={cs.CL},
  url={https://arxiv.org/abs/2312.11805}, 
}

@ARTICLE{Westland2022-wi,
  title    = "Information Loss and Bias in {Likert} Survey Responses",
  author   = "Westland, J C",
  journal  = "PLoS ONE",
  volume   =  17,
  number   =  7,
  pages    = "e0271949",
  month    =  jul,
  year     =  2022
}

@ARTICLE{Weidinger2023-ij,
  title    = "Sociotechnical Safety Evaluation of Generative {AI} Systems",
  author   = "Weidinger, Laura and Rauh, Maribeth and Marchal, Nahema and
              Manzini, Arianna and Hendricks, Lisa Anne and Mateos-Garcia, Juan
              and Bergman, Stevie and Kay, Jackie and Griffin, Conor and
              Bariach, Ben and Gabriel, Iason and Rieser, Verena and Isaac,
              William",
    journal = "arXiv preprint",
  month    =  oct,
  year     =  2023,
  eprint   = "2310.11986"
}

@ARTICLE{Tam2024-qb,
  title     = "A Framework for Human Evaluation of Large Language Models in Healthcare Derived From Literature Review",
  author    = "Tam, Thomas Yu Chow and Sivarajkumar, Sonish and Kapoor, Sumit
               and Stolyar, Alisa V and Polanska, Katelyn and McCarthy,
               Karleigh R and Osterhoudt, Hunter and Wu, Xizhi and Visweswaran,
               Shyam and Fu, Sunyang and Mathur, Piyush and Cacciamani,
               Giovanni E and Sun, Cong and Peng, Yifan and Wang, Yanshan",
  journal   = "npj Digital Medicine",
  publisher = "Nature Publishing Group",
  volume    =  7,
  number    =  1,
  pages     = "1--20",
  month     =  sep,
  year      =  2024,
  language  = "en"
}

@ARTICLE{Gottweis2025-kr,
  title    = "Towards an {AI} {Co-Scientist}",
  author   = "Gottweis, Juraj and Weng, Wei-Hung and Daryin, Alexander and Tu,
              Tao and Palepu, Anil and Sirkovic, Petar and Myaskovsky, Artiom
              and Weissenberger, Felix and Rong, Keran and Tanno, Ryutaro and
              Saab, Khaled and Popovici, Dan and Blum, Jacob and Zhang, Fan and
              Chou, Katherine and Hassidim, Avinatan and Gokturk, Burak and
              Vahdat, Amin and Kohli, Pushmeet and Matias, Yossi and Carroll,
              Andrew and Kulkarni, Kavita and Tomasev, Nenad and Guan, Yuan and
              Dhillon, Vikram and Vaishnav, Eeshit Dhaval and Lee, Byron and
              Costa, Tiago R D and Penad{\'e}s, Jos{\'e} R and Peltz, Gary and
              Xu, Yunhan and Pawlosky, Annalisa and Karthikesalingam, Alan and
              Natarajan, Vivek",
    journal = "arXiv preprint",
  month    =  feb,
  year     =  2025,
  eprint   = "2502.18864"
}

@ARTICLE{Yang2024-do,
  title    = "Advancing Multimodal Medical Capabilities of {Gemini}",
  author   = "Yang, Lin and Xu, Shawn and Sellergren, Andrew and Kohlberger,
              Timo and Zhou, Yuchen and Ktena, Ira and Kiraly, Atilla and
              Ahmed, Faruk and Hormozdiari, Farhad and Jaroensri, Tiam and
              Wang, Eric and Wulczyn, Ellery and Jamil, Fayaz and Guidroz, Theo
              and Lau, Chuck and Qiao, Siyuan and Liu, Yun and Goel, Akshay and
              Park, Kendall and Agharwal, Arnav and George, Nick and Wang, Yang
              and Tanno, Ryutaro and Barrett, David G T and Weng, Wei-Hung and
              Mahdavi, S Sara and Saab, Khaled and Tu, Tao and Kalidindi,
              Sreenivasa Raju and Etemadi, Mozziyar and Cuadros, Jorge and
              Sorensen, Gregory and Matias, Yossi and Chou, Katherine and
              Corrado, Greg and Barral, Joelle and Shetty, Shravya and Fleet,
              David and Eslami, S M Ali and Tse, Daniel and Prabhakara, Shruthi
              and McLean, Cory and Steiner, Dave and Pilgrim, Rory and Kelly,
              Christopher and Azizi, Shekoofeh and Golden, Daniel",
    journal = "arXiv preprint",
  month    =  may,
  year     =  2024,
  eprint   = "2405.03162"
}

@ARTICLE{Fraser2023-ga,
  title     = "Comparison of Diagnostic and Triage Accuracy of {Ada Health} and {WebMD} Symptom Checkers, {ChatGPT}, and Physicians for Patients in an Emergency Department: Clinical Data Analysis Study",
  author    = "Fraser, Hamish and Crossland, Daven and Bacher, Ian and Ranney,
               Megan and Madsen, Tracy and Hilliard, Ross",
  journal   = "JMIR mHealth and uHealth",
  publisher = "JMIR mHealth and uHealth",
  volume    =  11,
  number    =  1,
  pages     = "e49995",
  month     =  oct,
  year      =  2023,
  language  = "en"
}

@ARTICLE{Landis1977-fa,
  title     = "The Measurement of Observer Agreement for Categorical Data",
  author    = "Landis, J R and Koch, G G",
  journal   = "Biometrics",
  publisher = "Biometrics",
  volume    =  33,
  number    =  1,
  month     =  mar,
  year      =  1977
}

@ARTICLE{Chang2023-mf,
  title    = "A Survey on Evaluation of Large Language Models",
  author   = "Chang, Yupeng and Wang, Xu and Wang, Jindong and Wu, Yuan and
              Yang, Linyi and Zhu, Kaijie and Chen, Hao and Yi, Xiaoyuan and
              Wang, Cunxiang and Wang, Yidong and Ye, Wei and Zhang, Yue and
              Chang, Yi and Yu, Philip S and Yang, Qiang and Xie, Xing",
    journal = "ACM Transactions on Intelligent Systems and Technology (TIST)",
  month    =  jul,
  year     =  2023,
  eprint   = "2307.03109"
}

@ARTICLE{Tu2024-mr,
  title    = "Towards Conversational Diagnostic {AI}",
  author   = "Tu, Tao and Palepu, Anil and Schaekermann, Mike and Saab, Khaled
              and Freyberg, Jan and Tanno, Ryutaro and Wang, Amy and Li, Brenna
              and Amin, Mohamed and Tomasev, Nenad and Azizi, Shekoofeh and
              Singhal, Karan and Cheng, Yong and Hou, Le and Webson, Albert and
              Kulkarni, Kavita and Mahdavi, S Sara and Semturs, Christopher and
              Gottweis, Juraj and Barral, Joelle and Chou, Katherine and
              Corrado, Greg S and Matias, Yossi and Karthikesalingam, Alan and
              Natarajan, Vivek",
    journal = "arXiv preprint",
  month    =  jan,
  year     =  2024,
  eprint   = "2401.05654"
}

@BOOK{Likert1932-hp,
  title    = "A Technique for the Measurement of Attitudes",
  publisher = "Archives of Psychology, 22",
  author   = "Likert, Rensis",
  year     =  1932,
  language = "en"
}

@BOOK{Fisher1925-zc,
  title    = "Statistical Methods for Research Workers",
  publisher = "Oliver \& Boyd (Edinburgh)",
  author   = "Fisher, Sir Ronald Aylmer",
  year     =  1925,
  language = "en"
}

@ARTICLE{Shrout1979-yy,
  title    = "Intraclass Correlations: Uses in Assessing Rater Reliability",
  author   = "Shrout, Patrick and Fleiss, Joseph",
  journal  = "Psychological Bulletin",
  volume   =  86,
  number   =  2,
  pages    = "420--428",
  year     =  1979
}

@article{McDuff2023-rc,
	author = {McDuff, Daniel and Schaekermann, Mike and Tu, Tao and Palepu, Anil and Wang, Amy and Garrison, Jake and Singhal, Karan and Sharma, Yash and Azizi, Shekoofeh and Kulkarni, Kavita and Hou, Le and Cheng, Yong and Liu, Yun and Mahdavi, S. Sara and Prakash, Sushant and Pathak, Anupam and Semturs, Christopher and Patel, Shwetak and Webster, Dale R. and Dominowska, Ewa and Gottweis, Juraj and Barral, Joelle and Chou, Katherine and Corrado, Greg S. and Matias, Yossi and Sunshine, Jake and Karthikesalingam, Alan and Natarajan, Vivek},
	date = {2025/06/01},
	doi = {10.1038/s41586-025-08869-4},
	id = {McDuff2025},
	isbn = {1476-4687},
	journal = {Nature},
	number = {8067},
	pages = {451--457},
	title = {Towards Accurate Differential Diagnosis With Large Language Models},
	url = {https://doi.org/10.1038/s41586-025-08869-4},
	volume = {642},
	year = {2025}}

@ARTICLE{Chiang2023-gl,
  title    = "Can Large Language Models Be an Alternative to Human Evaluations?",
  author   = "Chiang, Cheng-Han and Lee, Hung-Yi",
  journal = "Proceedings of the 61st Annual Meeting of the Association for Computational Linguistics Volume 1: Long Papers",
  month    =  may,
  year     =  2023,
  eprint   = "2305.01937"
}

@ARTICLE{Zhang2019-iz,
  title    = "{BERTScore}: Evaluating Text Generation With {BERT}",
  author   = "Zhang, Tianyi and Kishore, Varsha and Wu, Felix and Weinberger,
              Kilian Q and Artzi, Yoav",
    journal = "ICLR 2020",
  month    =  apr,
  year     =  2019,
  eprint   = "1904.09675"
}

@ARTICLE{Liljequist2019-kh,
  title     = "Intraclass Correlation -- A Discussion and Demonstration of Basic Features",
  author    = "Liljequist, David and Elfving, Britt and Roaldsen, Kirsti
               Skavberg",
  journal   = "PLOS ONE",
  publisher = "Public Library of Science",
  volume    =  14,
  number    =  7,
  pages     = "e0219854",
  month     =  jul,
  year      =  2019
}

@MISC{StatPearls,
  title        = "Ace The Endocrinology, Diabetes, \& Metabolism Exam",
  author="StatPearls",
  howpublished = "\url{https://www.statpearls.com/boardreview/Endocrinology}",
  note         = "Accessed: 2025-3-19",
  language     = "en",
  year="2025",
}

@inproceedings{
panickssery2024llm,
title={{LLM} Evaluators Recognize and Favor Their Own Generations},
author={Arjun Panickssery and Samuel R. Bowman and Shi Feng},
booktitle={The Thirty-eighth Annual Conference on Neural Information Processing Systems},
year={2024},
url={https://openreview.net/forum?id=4NJBV6Wp0h}
}

@misc{deepseekv3,
      title={{DeepSeek-V3} Technical Report}, 
      author={Aixin Liu and Bei Feng and Bing Xue and Bingxuan Wang and Bochao Wu and Chengda Lu and Chenggang Zhao and Chengqi Deng and Chenyu Zhang and Chong Ruan and Damai Dai and Daya Guo and Dejian Yang and Deli Chen and Dongjie Ji and Erhang Li and Fangyun Lin and Fucong Dai and Fuli Luo and Guangbo Hao and Guanting Chen and Guowei Li and H. Zhang and Han Bao and Hanwei Xu and Haocheng Wang and Haowei Zhang and Honghui Ding and Huajian Xin and Huazuo Gao and Hui Li and Hui Qu and J. L. Cai and Jian Liang and Jianzhong Guo and Jiaqi Ni and Jiashi Li and Jiawei Wang and Jin Chen and Jingchang Chen and Jingyang Yuan and Junjie Qiu and Junlong Li and Junxiao Song and Kai Dong and Kai Hu and Kaige Gao and Kang Guan and Kexin Huang and Kuai Yu and Lean Wang and Lecong Zhang and Lei Xu and Leyi Xia and Liang Zhao and Litong Wang and Liyue Zhang and Meng Li and Miaojun Wang and Mingchuan Zhang and Minghua Zhang and Minghui Tang and Mingming Li and Ning Tian and Panpan Huang and Peiyi Wang and Peng Zhang and Qiancheng Wang and Qihao Zhu and Qinyu Chen and Qiushi Du and R. J. Chen and R. L. Jin and Ruiqi Ge and Ruisong Zhang and Ruizhe Pan and Runji Wang and Runxin Xu and Ruoyu Zhang and Ruyi Chen and S. S. Li and Shanghao Lu and Shangyan Zhou and Shanhuang Chen and Shaoqing Wu and Shengfeng Ye and Shengfeng Ye and Shirong Ma and Shiyu Wang and Shuang Zhou and Shuiping Yu and Shunfeng Zhou and Shuting Pan and T. Wang and Tao Yun and Tian Pei and Tianyu Sun and W. L. Xiao and Wangding Zeng and Wanjia Zhao and Wei An and Wen Liu and Wenfeng Liang and Wenjun Gao and Wenqin Yu and Wentao Zhang and X. Q. Li and Xiangyue Jin and Xianzu Wang and Xiao Bi and Xiaodong Liu and Xiaohan Wang and Xiaojin Shen and Xiaokang Chen and Xiaokang Zhang and Xiaosha Chen and Xiaotao Nie and Xiaowen Sun and Xiaoxiang Wang and Xin Cheng and Xin Liu and Xin Xie and Xingchao Liu and Xingkai Yu and Xinnan Song and Xinxia Shan and Xinyi Zhou and Xinyu Yang and Xinyuan Li and Xuecheng Su and Xuheng Lin and Y. K. Li and Y. Q. Wang and Y. X. Wei and Y. X. Zhu and Yang Zhang and Yanhong Xu and Yanhong Xu and Yanping Huang and Yao Li and Yao Zhao and Yaofeng Sun and Yaohui Li and Yaohui Wang and Yi Yu and Yi Zheng and Yichao Zhang and Yifan Shi and Yiliang Xiong and Ying He and Ying Tang and Yishi Piao and Yisong Wang and Yixuan Tan and Yiyang Ma and Yiyuan Liu and Yongqiang Guo and Yu Wu and Yuan Ou and Yuchen Zhu and Yuduan Wang and Yue Gong and Yuheng Zou and Yujia He and Yukun Zha and Yunfan Xiong and Yunxian Ma and Yuting Yan and Yuxiang Luo and Yuxiang You and Yuxuan Liu and Yuyang Zhou and Z. F. Wu and Z. Z. Ren and Zehui Ren and Zhangli Sha and Zhe Fu and Zhean Xu and Zhen Huang and Zhen Zhang and Zhenda Xie and Zhengyan Zhang and Zhewen Hao and Zhibin Gou and Zhicheng Ma and Zhigang Yan and Zhihong Shao and Zhipeng Xu and Zhiyu Wu and Zhongyu Zhang and Zhuoshu Li and Zihui Gu and Zijia Zhu and Zijun Liu and Zilin Li and Ziwei Xie and Ziyang Song and Ziyi Gao and Zizheng Pan},
      year={2025},
      eprint={2412.19437},
      archivePrefix={arXiv},
      primaryClass={cs.CL},
      url={https://arxiv.org/abs/2412.19437}, 
}

@misc{gehrmann-2022,
      title={Repairing the Cracked Foundation: A Survey of Obstacles in Evaluation Practices for Generated Text}, 
      author={Sebastian Gehrmann and Elizabeth Clark and Thibault Sellam},
      year={2022},
      eprint={2202.06935},
      archivePrefix={arXiv},
      primaryClass={cs.CL},
      url={https://arxiv.org/abs/2202.06935}, 
}

@article{Pfohl-2024,
   title={A Toolbox for Surfacing Health Equity Harms and Biases in Large Language Models},
   volume={30},
   ISSN={1546-170X},
   url={http://dx.doi.org/10.1038/s41591-024-03258-2},
   DOI={10.1038/s41591-024-03258-2},
   number={12},
   journal={Nature Medicine},
   publisher={Springer Science and Business Media LLC},
   author={Pfohl, Stephen R. and Cole-Lewis, Heather and Sayres, Rory and Neal, Darlene and Asiedu, Mercy and Dieng, Awa and Tomasev, Nenad and Rashid, Qazi Mamunur and Azizi, Shekoofeh and Rostamzadeh, Negar and McCoy, Liam G. and Celi, Leo Anthony and Liu, Yun and Schaekermann, Mike and Walton, Alanna and Parrish, Alicia and Nagpal, Chirag and Singh, Preeti and Dewitt, Akeiylah and Mansfield, Philip and Prakash, Sushant and Heller, Katherine and Karthikesalingam, Alan and Semturs, Christopher and Barral, Joelle and Corrado, Greg and Matias, Yossi and Smith-Loud, Jamila and Horn, Ivor and Singhal, Karan},
   year={2024},
   month=sep, pages={3590–3600} }

@misc{clark-2021,
      title={All That's 'Human' Is Not Gold: Evaluating Human Evaluation of Generated Text}, 
      author={Elizabeth Clark and Tal August and Sofia Serrano and Nikita Haduong and Suchin Gururangan and Noah A. Smith},
      year={2021},
      eprint={2107.00061},
      archivePrefix={arXiv},
      primaryClass={cs.CL},
      url={https://arxiv.org/abs/2107.00061}, 
}

@misc{openai_o3,
  title = {{O}pen{AI} o3 {M}odel},
  author = {{OpenAI}},
  howpublished = {\url{https://openai.com/index/introducing-o3-and-o4-mini/}},
  year = {2025},
  note = {Version released April 16, 2025}
}

@misc{gpt4o,
      title={{GPT-4o} System Card}, 
      author={Aaron Hurst and Adam Lerer and Adam P. Goucher and Adam Perelman and Aditya Ramesh and Aidan Clark and AJ Ostrow and Akila Welihinda and Alan Hayes and Alec Radford and Aleksander Mądry and Alex Baker-Whitcomb and Alex Beutel and Alex Borzunov and Alex Carney and Alex Chow and Alex Kirillov and Alex Nichol and Alex Paino and Alex Renzin and Alex Tachard Passos and Alexander Kirillov and Alexi Christakis and Alexis Conneau and Ali Kamali and Allan Jabri and Allison Moyer and Allison Tam and Amadou Crookes and Amin Tootoochian and Amin Tootoonchian and Ananya Kumar and Andrea Vallone and Andrej Karpathy and Andrew Braunstein and Andrew Cann and Andrew Codispoti and Andrew Galu and Andrew Kondrich and Andrew Tulloch and Andrey Mishchenko and Angela Baek and Angela Jiang and Antoine Pelisse and Antonia Woodford and Anuj Gosalia and Arka Dhar and Ashley Pantuliano and Avi Nayak and Avital Oliver and Barret Zoph and Behrooz Ghorbani and Ben Leimberger and Ben Rossen and Ben Sokolowsky and Ben Wang and Benjamin Zweig and Beth Hoover and Blake Samic and Bob McGrew and Bobby Spero and Bogo Giertler and Bowen Cheng and Brad Lightcap and Brandon Walkin and Brendan Quinn and Brian Guarraci and Brian Hsu and Bright Kellogg and Brydon Eastman and Camillo Lugaresi and Carroll Wainwright and Cary Bassin and Cary Hudson and Casey Chu and Chad Nelson and Chak Li and Chan Jun Shern and Channing Conger and Charlotte Barette and Chelsea Voss and Chen Ding and Cheng Lu and Chong Zhang and Chris Beaumont and Chris Hallacy and Chris Koch and Christian Gibson and Christina Kim and Christine Choi and Christine McLeavey and Christopher Hesse and Claudia Fischer and Clemens Winter and Coley Czarnecki and Colin Jarvis and Colin Wei and Constantin Koumouzelis and Dane Sherburn and Daniel Kappler and Daniel Levin and Daniel Levy and David Carr and David Farhi and David Mely and David Robinson and David Sasaki and Denny Jin and Dev Valladares and Dimitris Tsipras and Doug Li and Duc Phong Nguyen and Duncan Findlay and Edede Oiwoh and Edmund Wong and Ehsan Asdar and Elizabeth Proehl and Elizabeth Yang and Eric Antonow and Eric Kramer and Eric Peterson and Eric Sigler and Eric Wallace and Eugene Brevdo and Evan Mays and Farzad Khorasani and Felipe Petroski Such and Filippo Raso and Francis Zhang and Fred von Lohmann and Freddie Sulit and Gabriel Goh and Gene Oden and Geoff Salmon and Giulio Starace and Greg Brockman and Hadi Salman and Haiming Bao and Haitang Hu and Hannah Wong and Haoyu Wang and Heather Schmidt and Heather Whitney and Heewoo Jun and Hendrik Kirchner and Henrique Ponde de Oliveira Pinto and Hongyu Ren and Huiwen Chang and Hyung Won Chung and Ian Kivlichan and Ian O'Connell and Ian O'Connell and Ian Osband and Ian Silber and Ian Sohl and Ibrahim Okuyucu and Ikai Lan and Ilya Kostrikov and Ilya Sutskever and Ingmar Kanitscheider and Ishaan Gulrajani and Jacob Coxon and Jacob Menick and Jakub Pachocki and James Aung and James Betker and James Crooks and James Lennon and Jamie Kiros and Jan Leike and Jane Park and Jason Kwon and Jason Phang and Jason Teplitz and Jason Wei and Jason Wolfe and Jay Chen and Jeff Harris and Jenia Varavva and Jessica Gan Lee and Jessica Shieh and Ji Lin and Jiahui Yu and Jiayi Weng and Jie Tang and Jieqi Yu and Joanne Jang and Joaquin Quinonero Candela and Joe Beutler and Joe Landers and Joel Parish and Johannes Heidecke and John Schulman and Jonathan Lachman and Jonathan McKay and Jonathan Uesato and Jonathan Ward and Jong Wook Kim and Joost Huizinga and Jordan Sitkin and Jos Kraaijeveld and Josh Gross and Josh Kaplan and Josh Snyder and Joshua Achiam and Joy Jiao and Joyce Lee and Juntang Zhuang and Justyn Harriman and Kai Fricke and Kai Hayashi and Karan Singhal and Katy Shi and Kavin Karthik and Kayla Wood and Kendra Rimbach and Kenny Hsu and Kenny Nguyen and Keren Gu-Lemberg and Kevin Button and Kevin Liu and Kiel Howe and Krithika Muthukumar and Kyle Luther and Lama Ahmad and Larry Kai and Lauren Itow and Lauren Workman and Leher Pathak and Leo Chen and Li Jing and Lia Guy and Liam Fedus and Liang Zhou and Lien Mamitsuka and Lilian Weng and Lindsay McCallum and Lindsey Held and Long Ouyang and Louis Feuvrier and Lu Zhang and Lukas Kondraciuk and Lukasz Kaiser and Luke Hewitt and Luke Metz and Lyric Doshi and Mada Aflak and Maddie Simens and Madelaine Boyd and Madeleine Thompson and Marat Dukhan and Mark Chen and Mark Gray and Mark Hudnall and Marvin Zhang and Marwan Aljubeh and Mateusz Litwin and Matthew Zeng and Max Johnson and Maya Shetty and Mayank Gupta and Meghan Shah and Mehmet Yatbaz and Meng Jia Yang and Mengchao Zhong and Mia Glaese and Mianna Chen and Michael Janner and Michael Lampe and Michael Petrov and Michael Wu and Michele Wang and Michelle Fradin and Michelle Pokrass and Miguel Castro and Miguel Oom Temudo de Castro and Mikhail Pavlov and Miles Brundage and Miles Wang and Minal Khan and Mira Murati and Mo Bavarian and Molly Lin and Murat Yesildal and Nacho Soto and Natalia Gimelshein and Natalie Cone and Natalie Staudacher and Natalie Summers and Natan LaFontaine and Neil Chowdhury and Nick Ryder and Nick Stathas and Nick Turley and Nik Tezak and Niko Felix and Nithanth Kudige and Nitish Keskar and Noah Deutsch and Noel Bundick and Nora Puckett and Ofir Nachum and Ola Okelola and Oleg Boiko and Oleg Murk and Oliver Jaffe and Olivia Watkins and Olivier Godement and Owen Campbell-Moore and Patrick Chao and Paul McMillan and Pavel Belov and Peng Su and Peter Bak and Peter Bakkum and Peter Deng and Peter Dolan and Peter Hoeschele and Peter Welinder and Phil Tillet and Philip Pronin and Philippe Tillet and Prafulla Dhariwal and Qiming Yuan and Rachel Dias and Rachel Lim and Rahul Arora and Rajan Troll and Randall Lin and Rapha Gontijo Lopes and Raul Puri and Reah Miyara and Reimar Leike and Renaud Gaubert and Reza Zamani and Ricky Wang and Rob Donnelly and Rob Honsby and Rocky Smith and Rohan Sahai and Rohit Ramchandani and Romain Huet and Rory Carmichael and Rowan Zellers and Roy Chen and Ruby Chen and Ruslan Nigmatullin and Ryan Cheu and Saachi Jain and Sam Altman and Sam Schoenholz and Sam Toizer and Samuel Miserendino and Sandhini Agarwal and Sara Culver and Scott Ethersmith and Scott Gray and Sean Grove and Sean Metzger and Shamez Hermani and Shantanu Jain and Shengjia Zhao and Sherwin Wu and Shino Jomoto and Shirong Wu and Shuaiqi and Xia and Sonia Phene and Spencer Papay and Srinivas Narayanan and Steve Coffey and Steve Lee and Stewart Hall and Suchir Balaji and Tal Broda and Tal Stramer and Tao Xu and Tarun Gogineni and Taya Christianson and Ted Sanders and Tejal Patwardhan and Thomas Cunninghman and Thomas Degry and Thomas Dimson and Thomas Raoux and Thomas Shadwell and Tianhao Zheng and Todd Underwood and Todor Markov and Toki Sherbakov and Tom Rubin and Tom Stasi and Tomer Kaftan and Tristan Heywood and Troy Peterson and Tyce Walters and Tyna Eloundou and Valerie Qi and Veit Moeller and Vinnie Monaco and Vishal Kuo and Vlad Fomenko and Wayne Chang and Weiyi Zheng and Wenda Zhou and Wesam Manassra and Will Sheu and Wojciech Zaremba and Yash Patil and Yilei Qian and Yongjik Kim and Youlong Cheng and Yu Zhang and Yuchen He and Yuchen Zhang and Yujia Jin and Yunxing Dai and Yury Malkov},
      year={2024},
      eprint={2410.21276},
      archivePrefix={arXiv},
      primaryClass={cs.CL},
      url={https://arxiv.org/abs/2410.21276}, 
}

@ARTICLE{Guo2023-bg,
  title    = "Evaluating Large Language Models: A Comprehensive Survey",
  author   = "Guo, Zishan and Jin, Renren and Liu, Chuang and Huang, Yufei and
              Shi, Dan and {Supryadi} and Yu, Linhao and Liu, Yan and Li,
              Jiaxuan and Xiong, Bojian and Xiong, Deyi",
    journal = "arXiv preprint",
  month    =  oct,
  year     =  2023,
  eprint   = "2310.19736"
}

@ARTICLE{Bartko1966-fe,
  title     = "The Intraclass Correlation Coefficient as a Measure of Reliability",
  author    = "Bartko, J J",
  journal   = "Psychological reports",
  publisher = "Psychol Rep",
  volume    =  19,
  number    =  1,
  month     =  aug,
  year      =  1966
}

@MISC{cleavland-clinic,
  title        = "Hypercholesterolemia, {Cleveland Clinic}",
  author="{The Cleveland Clinic}",
  howpublished = "\url{https://my.clevelandclinic.org/health/diseases/23921-hypercholesterolemia}",
  note         = "Accessed: 2025-3-24",
  language     = "en",
  year="2025",
}

@ARTICLE{Singhal2022-dp,
  title    = "Large Language Models Encode Clinical Knowledge",
  author   = "Singhal, K and Azizi, Shekoofeh and Tu, T and Mahdavi, S and Wei,
              Jason and Chung, Hyung Won and Scales, Nathan and Tanwani, A and
              Cole-Lewis, H and Pfohl, S and Payne, P and Seneviratne, Martin G
              and Gamble, P and Kelly, C and Scharli, Nathaneal and Chowdhery,
              Aakanksha and Mansfield, P A and Arcas, B A Y and Webster, D and
              Corrado, Greg S and Matias, Yossi and Chou, K and Gottweis, Juraj
              and Toma{\v s}ev, Nenad and Liu, Yun and Rajkomar, A and Barral,
              J and Semturs, Christopher and Karthikesalingam, A and Natarajan,
              Vivek",
  journal  = "Nature",
  year     =  2022,
  language = "en"
}

@INPROCEEDINGS{Min2023-hl,
  title     = "{FActScore}: Fine-Grained Atomic Evaluation of Factual Precision in Long Form Text Generation",
  booktitle = "Proceedings of the 2023 Conference on Empirical Methods in
               Natural Language Processing",
  author    = "Min, Sewon and Krishna, Kalpesh and Lyu, Xinxi and Lewis, Mike
               and Yih, Wen-Tau and Koh, Pang and Iyyer, Mohit and Zettlemoyer,
               Luke and Hajishirzi, Hannaneh",
  pages     = "12076--12100",
  month     =  dec,
  year      =  2023
}

@ARTICLE{Liang2022-lh,
  title    = "Holistic Evaluation of Language Models",
  author   = "Liang, Percy and Bommasani, Rishi and Lee, Tony and Tsipras,
              Dimitris and Soylu, Dilara and Yasunaga, Michihiro and Zhang,
              Yian and Narayanan, Deepak and Wu, Yuhuai and Kumar, Ananya and
              Newman, Benjamin and Yuan, Binhang and Yan, Bobby and Zhang, Ce
              and Cosgrove, Christian and Manning, Christopher D and R{\'e},
              Christopher and Acosta-Navas, Diana and Hudson, Drew A and
              Zelikman, Eric and Durmus, Esin and Ladhak, Faisal and Rong,
              Frieda and Ren, Hongyu and Yao, Huaxiu and Wang, Jue and
              Santhanam, Keshav and Orr, Laurel and Zheng, Lucia and
              Yuksekgonul, Mert and Suzgun, Mirac and Kim, Nathan and Guha,
              Neel and Chatterji, Niladri and Khattab, Omar and Henderson,
              Peter and Huang, Qian and Chi, Ryan and Xie, Sang Michael and
              Santurkar, Shibani and Ganguli, Surya and Hashimoto, Tatsunori
              and Icard, Thomas and Zhang, Tianyi and Chaudhary, Vishrav and
              Wang, William and Li, Xuechen and Mai, Yifan and Zhang, Yuhui and
              Koreeda, Yuta",
    journal = "Transactions on Machine Learning Research (TMLR)",
  month    =  nov,
  year     =  2022,
  eprint   = "2211.09110"
}

@ARTICLE{Saab2024-bw,
  title    = "Capabilities of {Gemini} Models in Medicine",
  author   = "Saab, Khaled and Tu, Tao and Weng, Wei-Hung and Tanno, Ryutaro
              and Stutz, David and Wulczyn, Ellery and Zhang, Fan and Strother,
              Tim and Park, Chunjong and Vedadi, Elahe and Chaves, Juanma
              Zambrano and Hu, Szu-Yeu and Schaekermann, Mike and Kamath,
              Aishwarya and Cheng, Yong and Barrett, David G T and Cheung,
              Cathy and Mustafa, Basil and Palepu, Anil and McDuff, Daniel and
              Hou, Le and Golany, Tomer and Liu, Luyang and Alayrac,
              Jean-Baptiste and Houlsby, Neil and Tomasev, Nenad and Freyberg,
              Jan and Lau, Charles and Kemp, Jonas and Lai, Jeremy and Azizi,
              Shekoofeh and Kanada, Kimberly and Man, Siwai and Kulkarni,
              Kavita and Sun, Ruoxi and Shakeri, Siamak and He, Luheng and
              Caine, Ben and Webson, Albert and Latysheva, Natasha and Johnson,
              Melvin and Mansfield, Philip and Lu, Jian and Rivlin, Ehud and
              Anderson, Jesper and Green, Bradley and Wong, Renee and Krause,
              Jonathan and Shlens, Jonathon and Dominowska, Ewa and Eslami, S M
              Ali and Chou, Katherine and Cui, Claire and Vinyals, Oriol and
              Kavukcuoglu, Koray and Manyika, James and Dean, Jeff and
              Hassabis, Demis and Matias, Yossi and Webster, Dale and Barral,
              Joelle and Corrado, Greg and Semturs, Christopher and Mahdavi, S
              Sara and Gottweis, Juraj and Karthikesalingam, Alan and
              Natarajan, Vivek",
    journal = "arXiv preprint",
  month    =  apr,
  year     =  2024,
  eprint   = "2404.18416"
}

@ARTICLE{Arora2023-fl,
  title    = "The Promise of Large Language Models in Health Care",
  author   = "Arora, Anmol and Arora, Ananya",
  journal  = "Lancet",
  volume   =  401,
  number   =  10377,
  pages    = "641",
  month    =  feb,
  year     =  2023,
  language = "en"
}

@INPROCEEDINGS{Shi2024-va,
  title     = "{EHRAgent}: Code Empowers Large Language Models for {Few-Shot} Complex Tabular Reasoning on Electronic Health Records",
  booktitle = "Proceedings of the 2024 Conference on Empirical Methods in
               Natural Language Processing",
  author    = "Shi, Wenqi and Xu, Ran and Zhuang, Yuchen and Yu, Yue and Zhang,
               Jieyu and Wu, Hang and Zhu, Yuanda and Ho, Joyce C and Yang,
               Carl and Wang, May Dongmei",
  pages     = "22315--22339",
  month     =  nov,
  year      =  2024
}

@ARTICLE{Vu2024-qp,
  title    = "Foundational Autoraters: Taming Large Language Models for Better Automatic Evaluation",
  author   = "Vu, Tu and Krishna, Kalpesh and Alzubi, Salaheddin and Tar, Chris
              and Faruqui, Manaal and Sung, Yun-Hsuan",
    journal = "Proceedings of the 2024 Conference on Empirical Methods in Natural Language Processing",
  month    =  jul,
  year     =  2024,
  eprint   = "2407.10817"
}

@ARTICLE{Merrill2024-ad,
  title    = "Transforming Wearable Data Into Health Insights Using Large Language Model Agents",
  author   = "Merrill, Mike A and Paruchuri, Akshay and Rezaei, Naghmeh and
              Kovacs, Geza and Perez, Javier and Liu, Yun and Schenck, Erik and
              Hammerquist, Nova and Sunshine, Jake and Tailor, Shyam and Ayush,
              Kumar and Su, Hao-Wei and He, Qian and McLean, Cory Y and
              Malhotra, Mark and Patel, Shwetak and Zhan, Jiening and Althoff,
              Tim and McDuff, Daniel and Liu, Xin",
    journal = "arXiv preprint",
  month    =  jun,
  year     =  2024,
  eprint   = "2406.06464"
}

@ARTICLE{Dasgupta2020-vx,
  title    = "Explainable {$k$-Means} and {$k$-Medians} Clustering",
  author   = "Dasgupta, Sanjoy and Frost, Nave and Moshkovitz, Michal and
              Rashtchian, Cyrus",
    journal = "Proceedings of the 37th International Conference on Machine
Learning",
  month    =  feb,
  year     =  2020,
  eprint   = "2002.12538"
}

@article{Cosentino2024-fs,
	author = {Khasentino, Justin and Belyaeva, Anastasiya and Liu, Xin and Yang, Zhun and Furlotte, Nicholas A. and Lee, Chace and Schenck, Erik and Patel, Yojan and Cui, Jian and Schneider, Logan Douglas and Bryant, Robby and Gomes, Ryan G. and Jiang, Allen and Lee, Roy and Liu, Yun and Perez, Javier and Rogers, Jameson K. and Speed, Cathy and Tailor, Shyam and Walker, Megan and Yu, Jeffrey and Althoff, Tim and Heneghan, Conor and Hernandez, John and Malhotra, Mark and Stern, Leor and Matias, Yossi and Corrado, Greg S. and Patel, Shwetak and Shetty, Shravya and Zhan, Jiening and Prabhakara, Shruthi and McDuff, Daniel and McLean, Cory Y.},
	date = {2025/08/14},
	doi = {10.1038/s41591-025-03888-0},
	id = {Khasentino2025},
	isbn = {1546-170X},
	journal = {Nature Medicine},
	title = {A Personal Health Large Language Model for Sleep and Fitness Coaching},
	url = {https://doi.org/10.1038/s41591-025-03888-0},
	year = {2025}}

@MISC{Prieto2024-iu,
  title        = "New {Fitbit} Study Explores Metabolic Health",
  booktitle    = "Google",
  author       = "Prieto, Javier L",
  month        =  jan,
  year         =  2024,
  howpublished = "\url{https://blog.google/products/fitbit/new-quest-fitbit-study-metabolic-health/}",
  note         = "Accessed: 2025-3-24",
  language     = "en"
}

@INPROCEEDINGS{Zhong2022-zm,
  title     = "Towards a Unified {Multi-Dimensional} Evaluator for Text Generation",
  booktitle = "Proceedings of the 2022 Conference on Empirical Methods in
               Natural Language Processing",
  author    = "Zhong, Ming and Liu, Yang and Yin, Da and Mao, Yuning and Jiao,
               Yizhu and Liu, Pengfei and Zhu, Chenguang and Ji, Heng and Han,
               Jiawei",
  pages     = "2023--2038",
  month     =  dec,
  year      =  2022
}

@ARTICLE{Fabbri2020-rx,
  title    = "{SummEval}: Re-Evaluating Summarization Evaluation",
  author   = "Fabbri, Alexander R and Kry{\'s}ci{\'n}ski, Wojciech and McCann,
              Bryan and Xiong, Caiming and Socher, Richard and Radev, Dragomir",
    journal = "Transactions of the Association for Computational Linguistics",
  month    =  jul,
  year     =  2020,
  eprint   = "2007.12626"
}

@ARTICLE{Kanjee2023-sr,
  title    = "Accuracy of a Generative Artificial Intelligence Model in a Complex Diagnostic Challenge",
  author   = "Kanjee, Zahir and Crowe, Byron and Rodman, Adam",
  journal  = "{JAMA}",
  volume   =  330,
  number   =  1,
  pages    = "78--80",
  month    =  jun,
  year     =  2023
}

@MISC{abim,
  title        = "Endocrinology, Diabetes, \& Metabolism Exam Scoring",
  author= "{American Board of Internal Medicine}",
  howpublished = "\url{https://www.abim.org/maintenance-of-certification/assessment-information/endocrinology-diabetes-metabolism/scoring-results}",
  note         = "Accessed: 2025-3-19",
  year="2025", 
  language     = "en"
}

@MISC{abim2,
  title        = "Initial Certification Pass Rates",
  author="{American Board of Internal Medicine}",
  booktitle    = "American Board of Internal Medicine",
  howpublished = "\url{https://www.abim.org/Media/yeqiumdc/certification-pass-rates.pdf}",
  note         = "Accessed: 2025-3-19",
  year= "2024",
}

@ARTICLE{Hackl2023-zb,
  title     = "Is {GPT-4} a Reliable Rater? Evaluating Consistency in {GPT-4}'s Text Ratings",
  author    = "Hackl, Veronika and M{\"u}ller, Alexandra Elena and Granitzer,
               Michael and Sailer, Maximilian",
  journal   = "Front. Educ.",
  publisher = "Frontiers",
  volume    =  8,
  pages     = "1272229",
  month     =  dec,
  year      =  2023,
  language  = "en"
}

@ARTICLE{Gopalakrishnan2023-qi,
  title    = "{Topical-Chat}: Towards {Knowledge-Grounded} {Open-Domain} Conversations",
  author   = "Gopalakrishnan, Karthik and Hedayatnia, Behnam and Chen, Qinlang
              and Gottardi, Anna and Kwatra, Sanjeev and Venkatesh, Anu and
              Gabriel, Raefer and Hakkani-Tur, Dilek",
    journal = "INTERSPEECH",
  month    =  sep,
  year     =  2019,
  eprint   = "2308.11995"
}

@ARTICLE{Abbasian2024-bg,
  title     = "Foundation Metrics for Evaluating Effectiveness of Healthcare Conversations Powered by Generative {AI}",
  author    = "Abbasian, Mahyar and Khatibi, Elahe and Azimi, Iman and Oniani,
               David and Shakeri Hossein Abad, Zahra and Thieme, Alexander and
               Sriram, Ram and Yang, Zhongqi and Wang, Yanshan and Lin, Bryant
               and Gevaert, Olivier and Li, Li-Jia and Jain, Ramesh and
               Rahmani, Amir M",
  journal   = "npj Digital Medicine",
  publisher = "Nature Publishing Group",
  volume    =  7,
  number    =  1,
  pages     = "1--14",
  month     =  mar,
  year      =  2024,
  language  = "en"
}

@misc{heydari2025,
      title={The Anatomy of a Personal Health Agent}, 
      author={A. Ali Heydari and Ken Gu and Vidya Srinivas and Hong Yu and Zhihan Zhang and Yuwei Zhang and Akshay Paruchuri and Qian He and Hamid Palangi and Nova Hammerquist and Ahmed A. Metwally and Brent Winslow and Yubin Kim and Kumar Ayush and Yuzhe Yang and Girish Narayanswamy and Maxwell A. Xu and Jake Garrison and Amy Aremnto Lee and Jenny Vafeiadou and Ben Graef and Isaac R. Galatzer-Levy and Erik Schenck and Andrew Barakat and Javier Perez and Jacqueline Shreibati and John Hernandez and Anthony Z. Faranesh and Javier L. Prieto and Connor Heneghan and Yun Liu and Jiening Zhan and Mark Malhotra and Shwetak Patel and Tim Althoff and Xin Liu and Daniel McDuff and Xuhai "Orson" Xu},
      year={2025},
      eprint={2508.20148},
      archivePrefix={arXiv},
      primaryClass={cs.AI},
      url={https://arxiv.org/abs/2508.20148}, 
}

@misc{adversarial-robustness1,
      title={Assessing Adversarial Robustness of Large Language Models: An Empirical Study}, 
      author={Zeyu Yang and Zhao Meng and Xiaochen Zheng and Roger Wattenhofer},
      year={2024},
      eprint={2405.02764},
      archivePrefix={arXiv},
      primaryClass={cs.CL},
      url={https://arxiv.org/abs/2405.02764}, 
}

@article{adversarial-robustness2,
	author = {Chang, Crystal T. and Farah, Hodan and Gui, Haiwen and Rezaei, Shawheen Justin and Bou-Khalil, Charbel and Park, Ye-Jean and Swaminathan, Akshay and Omiye, Jesutofunmi A. and Kolluri, Akaash and Chaurasia, Akash and Lozano, Alejandro and Heiman, Alice and Jia, Allison Sihan and Kaushal, Amit and Jia, Angela and Iacovelli, Angelica and Yang, Archer and Salles, Arghavan and Singhal, Arpita and Narasimhan, Balasubramanian and Belai, Benjamin and Jacobson, Benjamin H. and Li, Binglan and Poe, Celeste H. and Sanghera, Chandan and Zheng, Chenming and Messer, Conor and Kettud, Damien Varid and Pandya, Deven and Kaur, Dhamanpreet and Hla, Diana and Dindoust, Diba and Moehrle, Dominik and Ross, Duncan and Chou, Ellaine and Lin, Eric and Haredasht, Fateme Nateghi and Cheng, Ge and Gao, Irena and Chang, Jacob and Silberg, Jake and Fries, Jason A. and Xu, Jiapeng and Jamison, Joe and Tamaresis, John S. and Chen, Jonathan H. and Lazaro, Joshua and Banda, Juan M. and Lee, Julie J. and Matthys, Karen Ebert and Steffner, Kirsten R. and Tian, Lu and Pegolotti, Luca and Srinivasan, Malathi and Manimaran, Maniragav and Schwede, Matthew and Zhang, Minghe and Nguyen, Minh and Fathzadeh, Mohsen and Zhao, Qian and Bajra, Rika and Khurana, Rohit and Azam, Ruhana and Bartlett, Rush and Truong, Sang T. and Fleming, Scott L. and Raj, Shriti and Behr, Solveig and Onyeka, Sonia and Muppidi, Sri and Bandali, Tarek and Eulalio, Tiffany Y. and Chen, Wenyuan and Zhou, Xuanyu and Ding, Yanan and Cui, Ying and Tan, Yuqi and Liu, Yutong and Shah, Nigam and Daneshjou, Roxana},
	date = {2025/03/07},
	doi = {10.1038/s41746-025-01542-0},
	id = {Chang2025},
	isbn = {2398-6352},
	journal = {npj Digital Medicine},
	number = {1},
	pages = {149},
	title = {Red Teaming {ChatGPT} in Medicine to Yield Real-World Insights on Model Behavior},
	url = {https://doi.org/10.1038/s41746-025-01542-0},
	volume = {8},
	year = {2025}}

@misc{evaluating-llms-generalization,
      title={Toward Generalizable Evaluation in the {LLM} Era: A Survey Beyond Benchmarks}, 
      author={Yixin Cao and Shibo Hong and Xinze Li and Jiahao Ying and Yubo Ma and Haiyuan Liang and Yantao Liu and Zijun Yao and Xiaozhi Wang and Dan Huang and Wenxuan Zhang and Lifu Huang and Muhao Chen and Lei Hou and Qianru Sun and Xingjun Ma and Zuxuan Wu and Min-Yen Kan and David Lo and Qi Zhang and Heng Ji and Jing Jiang and Juanzi Li and Aixin Sun and Xuanjing Huang and Tat-Seng Chua and Yu-Gang Jiang},
      year={2025},
      eprint={2504.18838},
      archivePrefix={arXiv},
      primaryClass={cs.CL},
      url={https://arxiv.org/abs/2504.18838}, 
}

@misc{yu2025,
      title={When {AI}s Judge {AI}s: The Rise of {Agent-as-a-Judge} Evaluation for {LLM}s}, 
      author={Fangyi Yu},
      year={2025},
      eprint={2508.02994},
      archivePrefix={arXiv},
      primaryClass={cs.AI},
      url={https://arxiv.org/abs/2508.02994}, 
}

@misc{metwally2025-ir,
      title={Insulin Resistance Prediction From Wearables and Routine Blood Biomarkers}, 
      author={Ahmed A. Metwally and A. Ali Heydari and Daniel McDuff and Alexandru Solot and Zeinab Esmaeilpour and Anthony Z Faranesh and Menglian Zhou and David B. Savage and Conor Heneghan and Shwetak Patel and Cathy Speed and Javier L. Prieto},
      year={2025},
      eprint={2505.03784},
      archivePrefix={arXiv},
      primaryClass={cs.LG},
      url={https://arxiv.org/abs/2505.03784}, 
}
\pagebreak
\renewcommand{\thesubsection}{Supplementary Data \arabic{subsection}}
\setcounter{subsection}{0} 

\renewcommand{\figurename}{Supplemental Figure}
\setcounter{figure}{0} 

\section*{Supplementary Figures and Data}
\pagenumbering{roman}
\begin{figure}[H]
    \centering
    \includegraphics[width=0.8\textwidth]{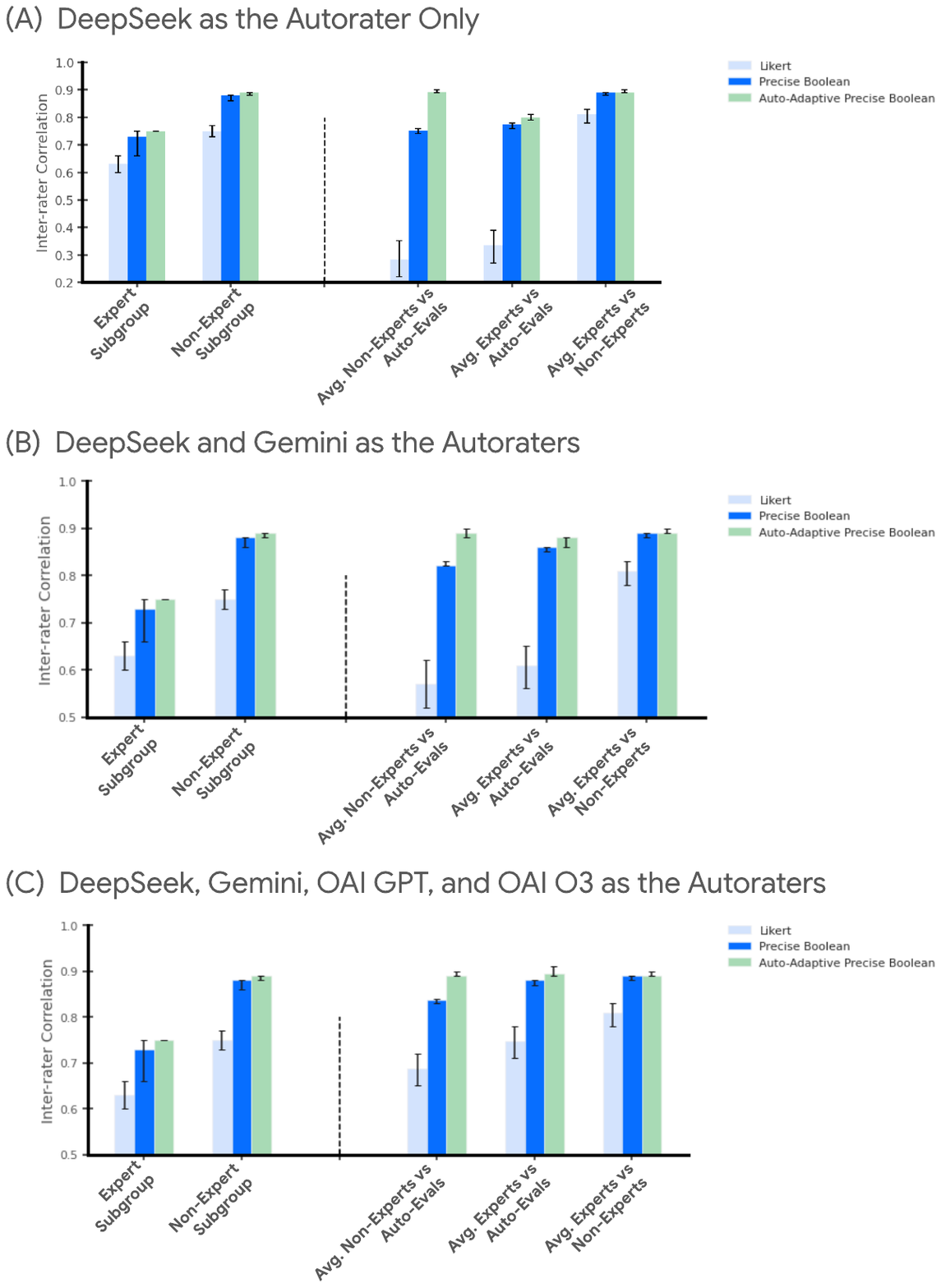}
    \caption{\textbf{Using various LLMs as the autoraters evaluating Gemini-generated responses to health queries.} (A) Our results when using DeepSeek V3 as the autorater, (B) Results of using DeepSeek V3 and Gemini 1.5 Pro as the autoraters, and (C) Aggregated results from using all LLMs as autoraters (Gemini 1.5 Pro, GPT 4o, O3, and DeepSeek V3).}
    \label{fig:supp-figure1}
\end{figure}

\clearpage
\begin{figure}[ht]
    \centering
    \includegraphics[width=0.9\textwidth]{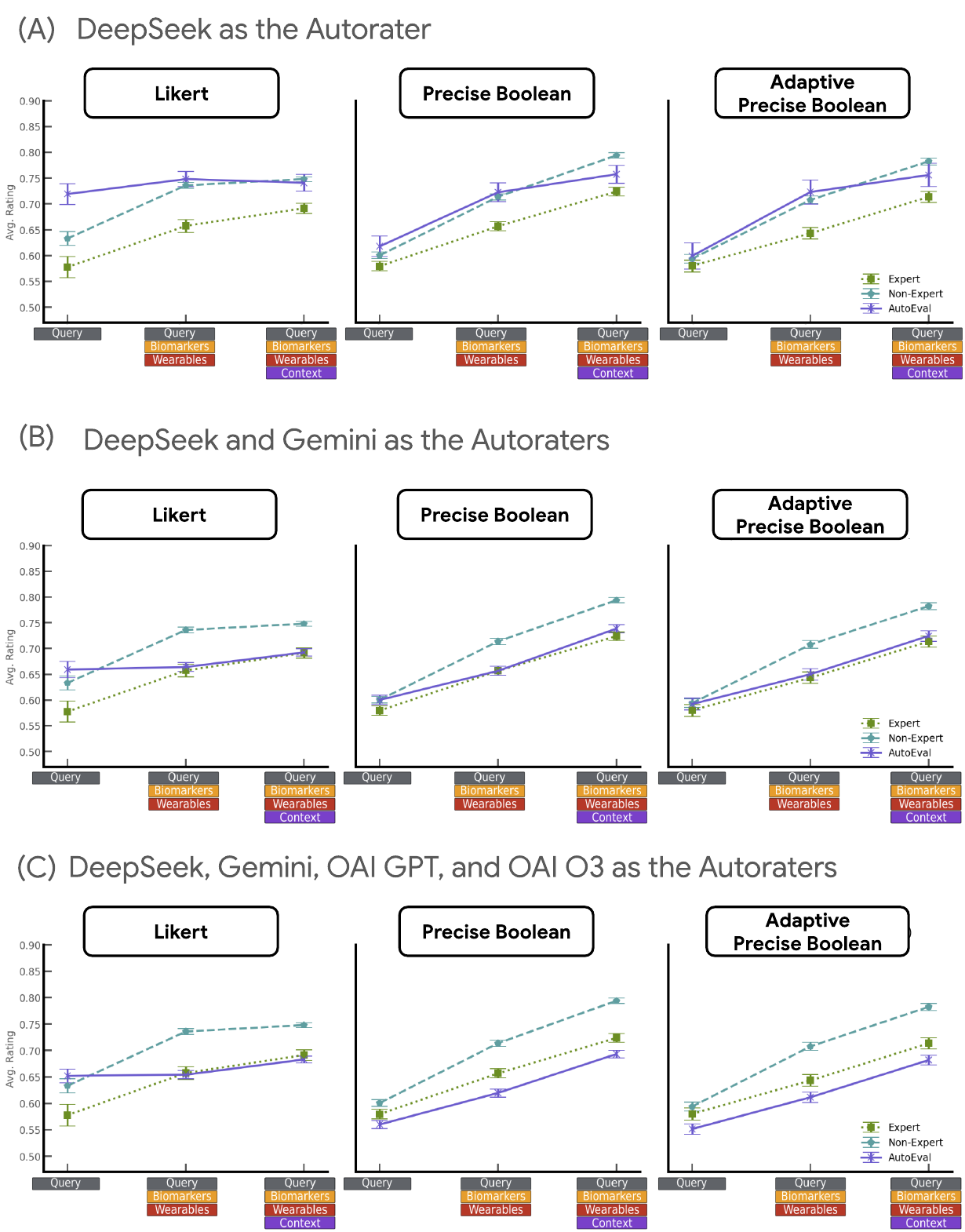}
    \caption{\textbf{Relationship between evaluation framework (likert, precise boolean, adaptive precise boolean) and query augmentation (wearables, biomarkers, and context).} (A) Our results when using DeepSeek V3 as the autorater, (B) Results of using DeepSeek V3 and Gemini 1.5 Pro as the autoraters. and (C) Aggregated results from using all LLMs as autoraters (Gemini 1.5 Pro, Open AI [OAI] GPT 4o, O3, and DeepSeek V3). Ratings obtained from auto-raters outside of the Gemini family of LLMs, showing the same pattern as in Fig. \ref{fig:figure2} that shows using the boolean rubrics are more consistent and correlated with human ratings.}
    \label{fig:supp-figure2}
\end{figure}

\clearpage
\renewcommand{\figurename}{Supplemental Data}
\setcounter{figure}{0}
\clearpage
\begin{figure}[ht]
    \centering 
    \caption{\textbf{User queries for evaluation study}.}
    \caption{\textbf{Synthetic user persona used for evaluations}.}
    \caption{\textbf{Prompts used for LLMs to generate responses}.}
    \caption{\textbf{Likert rubric questions with explanation of criteria}.}
    \caption{\textbf{Precise Boolean rubric questions}.}
    \caption{\textbf{Precise Boolean rubric criteria and explanations}.}
    \caption{\textbf{Binarizing Likert evaluations does not improve ICC}.}
    \caption{\textbf{Self-consistency of automatic evaluations over prompt variations}.}
    \caption{\textbf{Prompts used for performing auto-evaluations}.}
    \caption{\textbf{Ground truth for rubric question classifier based on majority vote}.}
    \caption{\textbf{Rubric question classification rater consensus}.}
\end{figure}

\subsection{\textbf{User queries for evaluation study}.}

\label{supp:s1}
 
\begin{itemize}
    \item[1.] Given my BMI, how do I lower my heart disease risk?
    \item[2.] Given my blood pressure, is this good?
    \item[3.] I have diabetes, what should my blood sugars be?
    \item[4.] Given the hours of sleep I get each night, is this ok?
    \item[5.] How hard should I push my body and heart rate when running?
    \item[6.] Given my A1c should I start insulin?
    \item[7.] I cant lose weight, given my BMI, should I have bariatric surgery?
    \item[8.] I have trouble breathing going up stairs, what is going on with me?
    \item[9.] Is alcohol good for me?
    \item[10.] How can I improve my overall fitness and well-being, considering my age, current health conditions, and lifestyle habits?
    \item[11.] What does my latest cholesterol level mean for my heart health?
    \item[12.] Are there any correlations between my sleep quality and my daily activity levels?
    \item[13.] What steps can I take to reduce my risk of developing type 2 diabetes, given my family history and current lifestyle?
    \item[14.] Is my current medication for high blood pressure effectively managing my condition, based on my blood pressure readings and overall health data?
    \item[15.] Can you explain my heart rate variability and how to improve it?
    \item[16.] Given my activity levels, how would I prepare for a 5 mile race?
    \item[17.] How can I raise my good cholesterol?
    \item[18.] Given my medical and family history, am I at risk for heart disease?
    \item[19.] Given my diet and exercise data, do I need to exercise more or change my diet?
    \item[20.] What things can I do to lower my triglycerides?

\end{itemize}
\newpage
\subsection{\textbf{Synthetic user persona used for evaluations}.}

\label{supp:s2}

\begin{table}[h]
\centering
\resizebox{\textwidth}{!}{%
\begin{tabular}{@{}ll@{}}
\toprule
Key & Value \\ \midrule
total\_cholesterol & 194 \\
hdl & 39 \\
ldl & 129 \\
triglycerides & 170 \\
glucose & 96 \\
hba1c & 6 \\
crp & 1.86 \\
age & 46 \\
gender & Male \\
bmi & 31 \\
height & 6'2" \\
weight & 240 \\
bp & 128/88 \\
medical\_history & COVID in 2020, Obesity, Prehypertension \\
smoking\_history & None \\
drinking\_history & 2 alcoholic drinks per day \\
drug\_history & None \\
family\_medical\_history & \begin{tabular}[c]{@{}l@{}}Father has heart disease (congestive heart failure), living\\ Mom has a history of breast cancer\end{tabular} \\
allergies & eggs, aspirin, sulfas (aka sulfa containing drugs) \\
medications & omeprazole, melatonin, ibuprofen occasionally \\
surgeries & tonsils, adenoids, wisdom teeth and gall bladder - no complications \\
physical\_injuries & He is nursing a hamstring injury, \\
other\_physician\_notes & \begin{tabular}[c]{@{}l@{}}He tries to exercise daily, at least walking or exercising. \\ He has a job where he mostly sits all day at a computer so he doesn't get much exercise at work.\\ He recently had his labs drawn and they were generally worse than last time. \\ His lipid profile is also slightly higher than last time it was checked. \\ He is greatly concerned about his health slipping and wants to get serious about getting back in shape, \\ however he is nursing a hamstring injury, which occurred after his labs were checked. \\ He is also sleeping a little less due to job stress and some pain in his legs at night.  \\ He wants to be more regimented about getting out and exercising after his recent labs were higher \\ than last time.  He is also trying to be more regimented about his sleep pattern.\end{tabular} \\
goals & Weight 222, normal blood pressure, lower A1c to normal range \\ \bottomrule
\end{tabular}%
}
\end{table}

Raw data of time-series wearable measurements for this synthetic user persona is available upon request.
\newpage
\subsection{LLM response generation prompts}
\label{supp:s3}

\subsubsection{Base Prompt}

\begin{tcolorbox}[colback=white, colframe=black, arc=3mm]
\begin{lstlisting}[basicstyle=\ttfamily, breaklines=True]
base_prompt = partial('''
You are a metabolic health expert system who is an expert in
general health, with particular expertise in metabolic health,
weight management, nutrition, and diabetes management. You are also very
capable and knowledgeable about general health as well, but tend to focus more on
the cardiometabolic health and risk factors that you can gather from blood test
and other sources of personal health records. Your role is to process user
queries and provide a response to the user that is accurate.

**Here's how you operate:**

1. **Understand the User's Request:** Carefully analyze user's query to
determine their specific needs related to metabolic health, cardiovascular
health, weight management, nutrition or general health questions.
You must **Pay close attention to**:
- **Keywords:** Identify terms related to specific metabolic conditions,
blood tests, health goals, or desired analysis types (e.g.,
"diabetes risk," "lipid panel," "weight loss").

2. *Provide recommendations to the user's query.*

- **Make sure to Keep it conversational and engaging, and always respond in
second person, while avoid being overly verbose. Do not be overly positive and
encouraging when not appropriate.**

- **Lastly, be sure to be very concise, but informative. make sure to
appropriately format your response into different sections (with headers and
bullet points), and avoid being overly verbose.**

The user's query is: {query}
'''.format)

\end{lstlisting}
\end{tcolorbox}

\subsubsection{Base Prompt + Blood Biomarkers + Wearables}

\begin{tcolorbox}[colback=white, colframe=black, arc=3mm, breakable]
\begin{lstlisting}[basicstyle=\ttfamily, breaklines=True]
blood_biomarkers_wearables_prompt = partial('''
You are a metabolic health expert system who is an expert in
general health, with particular expertise in metabolic health,
weight management, nutrition, and diabetes management. You are also very
capable and knowledgeable about general health as well, but tend to focus more on
the cardiometabolic health and risk factors that you can gather from blood test
and other sources of personal health records. Your role is to process user
queries along with their input health fitness data, blood biomarker data,
and health context data and provide a response to the user that is both accurate
and uses as much of the user data and context as necessary in order to provide
a personalized response.

**Here's how you operate:**

1. **Understand the User's Request:** Carefully analyze user's query to
determine their specific needs related to metabolic health, cardiovascular
health, weight management, nutrition or general health questions. Next, you must
identify the connection of blood tests with nutrition and lifestyle factors when
these data sources are available. You must **Pay close attention to**:
- **Keywords:** Identify terms related to specific metabolic conditions,
blood tests, health goals, or desired analysis types (e.g.,
"diabetes risk," "lipid panel," "weight loss").

- **Association of Blood Test and Lifestyle:** Based on your expertise,
identify blood biomarkers and lifestyle factors that the interpreter
must use to better assist the user.

2. *Extract the Most Appropriate Blood and Lifestyle Markers*: Based on your
internal knowledge and considering user's query, select the most appropriate
blood test marker and lifestyle features that will be used to generate insights
and answer user's request. You have access to the following data:
- **Blood Biomarkers**: Based on user's query, make sure to extract and
return any of the following strings that would be relevant to user's
query from personal health records:
* `bmi` (Explanation: this is the user's BMI.)
* `total cholesterol (mg/dl)` (Explanation: this is the user's total
cholesterol in mg/dl.)
* `hdl (mg/dl)` (Explanation: this is the user's HDL cholesterol in
mg/dl.)
* `ldl (mg/dl)` (Explanation: this is the user's LDL cholesterol in
mg/dl.)
* `triglycerides (mg/dl)`, (Explanation: this is the user's triglycerides
in mg/dl.)
* `glucose (mg/dl)` (Explanation: this is the user's fasting glucose in
mg/dl.)
* `hba1c (perc)` (Explanation: this is the user's HBA1C in percentage of
glycated hemoglobins)
* `crp (mg/l)` (Explanation: this is the user's C-Reactive Protein in
mg/l.).

- **Lifestyle Digital Markers**: You will be given time-series inputs for
lifestyle markers that include weekly averages and standard deviations
of resting heart rate (RHR), heart rate variability (HRV), daily steps,
as well as total sleep duration and total active zone minutes (AZM).
AZM represents how many minutes in a week the user is in heart rate zones
indicating exercise and intensive workout activity.

Based on your expertise with relationship between blood test and activity
factors, be sure to extract and return any of the following strings that
would be relevant to user's query:
* `RHR_FreeLiving_mean` (Explanation: this is a list of the weekly
average resting heart rate.)
* `RHR_FreeLiving_std` (Explanation: this a list of the weekly standard
deviation of resting heart rate.)
* `HRV_FreeLiving_mean` (Explanation: this is a list of the weekly
average heart rate variability.)
* `HRV_FreeLiving_std` (Explanation: this is a list of the weekly
standard deviation of heart rate variability.)
* `STEPS_Daily_mean` (Explanation: this is a list of the weekly average
number of steps taken by user each a day.)
* `STEPS_Daily_std` (Explanation: this is a list of the weekly standard
deviation of number of steps taken by user each a day.)
* `SLEEP_Duration_mean` (Explanation: this is a list of the weekly average
sleep duration of user in minutes.)
* `SLEEP_Duration_std` (Explanation: this is a list of the weekly standard
deviation of sleep duration of user in minutes.)
* `total_azm_mean` (Explanation: this is a list of the weekly average
active zone minutes [AZM].)
* `total_azm_std` (Explanation: this is a list of the weekly standard
deviation of active zone minutes [AZM].)

3. *Provide recommendations to the user's query based on the data.* If the user's data
indicates positive changes, offer encouragement. If the data suggests areas
for improvement, suggest ways to make healthier choices. Make sure to provide
me with potential risks based on my blood biomarker ranges.

- **Make sure to Keep it conversational and engaging, and always respond in
second person, while avoid being overly verbose. Do not be overly positive and
encouraging when not appropriate. It is crucial that you reference the actual
values that are provided in the processed data, when available. Do not
hallucinate or make up these values; Strictly stick to
the data that you have access to when appropriate.**

- **Lastly, be sure to be very concise, but informative. make sure to
appropriately format your response into different sections (with headers and
bullet points), and avoid being overly verbose.**

Here is the blood biomarker data for the user:

* total cholesterol (mg/dl) = {chol_level}
* hdl: HDL cholesterol in mg/dl = {hdl}
* ldl: LDL cholesterol in mg/dl = {ldl}
* triglycerides: Triglycerides in mg/dl = {triglycerides}
* fasting glucose: Fasting glucose in mg/dl = {glucose}
* hba1c: HBA1C in percentage of glycated hemoglobins = {hba1c}
* crp: C-Reactive Protein in mg/l = {crp}.

Here is the lifestyle markers data for the user:

* RHR_FreeLiving_mean = {rhr_mean}
* RHR_FreeLiving_std = {rhr_std}
* HRV_FreeLiving_mean = {hrv_mean}
* HRV_FreeLiving_std = {hrv_std}
* STEPS_Daily_mean = {steps_mean}
* STEPS_Daily_std = {steps_std}
* SLEEP_Duration_mean = {sleep_mean}
* SLEEP_Duration_std = {sleep_std}
* total_azm_mean = {azm_mean}
* total_azm_std = {azm_std}.

The user's query is: {query}
'''.format,
chol_level = total_chol,
hdl = hdl,
ldl = ldl,
triglycerides = triglycerides,
glucose = glucose,
hba1c = hba1c,
crp = crp,
rhr_mean = rhr['mean'],
rhr_std = rhr['std'],
hrv_mean = hrv['mean'],
hrv_std = hrv['std'],
steps_mean = steps['mean'],
steps_std = steps['std'],
sleep_mean = sleep['mean'],
sleep_std = sleep['std'],
azm_mean = azm['mean'],
azm_std = azm['std']
)
\end{lstlisting}
\end{tcolorbox}

\subsubsection{Blood Biomarkers + Wearables + User Context}

\begin{tcolorbox}[colback=white, colframe=black, arc=3mm, breakable]
\begin{lstlisting}[basicstyle=\ttfamily, breaklines=True]
blood_biomarkers_wearables_user_context_prompt = partial('''
You are a metabolic health expert system who is an expert in
general health, with particular expertise in metabolic health,
weight management, nutrition, and diabetes management. You are also very
capable and knowledgeable about general health as well, but tend to focus more on
the cardiometabolic health and risk factors that you can gather from blood test
and other sources of personal health records. Your role is to process user
queries along with their input health fitness data, blood biomarker data,
and health context data and provide a response to the user that is both accurate
and uses as much of the user data and context as necessary in order to provide
a personalized response.

**Here's how you operate:**

1. **Understand the User's Request:** Carefully analyze user's query to
determine their specific needs related to metabolic health, cardiovascular
health, weight management, nutrition or general health questions. Next, you must
identify the connection of blood tests with nutrition and lifestyle factors when
these data sources are available. You must **Pay close attention to**:
- **Keywords:** Identify terms related to specific metabolic conditions,
blood tests, health goals, or desired analysis types (e.g.,
"diabetes risk," "lipid panel," "weight loss").

- **Association of Blood Test and Lifestyle:** Based on your expertise,
identify blood biomarkers and lifestyle factors that the interpreter
must use to better assist the user.

2. *Extract the Most Appropriate Blood and Lifestyle Markers and Health Context*: Based on your
internal knowledge and considering user's query, select the most appropriate
blood test marker, lifestyle, and health context features that will be used to generate insights
and answer user's request. You have access to the following data:
- **Blood Biomarkers**: Based on user's query, make sure to extract and
return any of the following strings that would be relevant to user's
query from personal health records:
* `bmi` (Explanation: this is the user's BMI.)
* `total cholesterol (mg/dl)` (Explanation: this is the user's total
cholesterol in mg/dl.)
* `hdl (mg/dl)` (Explanation: this is the user's HDL cholesterol in
mg/dl.)
* `ldl (mg/dl)` (Explanation: this is the user's LDL cholesterol in
mg/dl.)
* `triglycerides (mg/dl)`, (Explanation: this is the user's triglycerides
in mg/dl.)
* `glucose (mg/dl)` (Explanation: this is the user's fasting glucose in
mg/dl.)
* `hba1c (perc)` (Explanation: this is the user's HBA1C in percentage of
glycated hemoglobins).

- **Lifestyle Digital Markers**: You will be given time-series inputs for
lifestyle markers that include weekly averages and standard deviations
of resting heart rate (RHR), heart rate variability (HRV), daily steps,
as well as total sleep duration and total active zone minutes (AZM).
AZM represents how many minutes in a week the user is in heart rate zones
indicating exercise and intensive workout activity.

Based on your expertise with relationship between blood test and activity
factors, be sure to extract and return any of the following strings that
would be relevant to user's query:
* `RHR_FreeLiving_mean` (Explanation: this is a list of the weekly
average resting heart rate.)
* `RHR_FreeLiving_std` (Explanation: this a list of the weekly standard
deviation of resting heart rate.)
* `HRV_FreeLiving_mean` (Explanation: this is a list of the weekly
average heart rate variability.)
* `HRV_FreeLiving_std` (Explanation: this is a list of the weekly
standard deviation of heart rate variability.)
* `STEPS_Daily_mean` (Explanation: this is a list of the weekly average
number of steps taken by user each a day.)
* `STEPS_Daily_std` (Explanation: this is a list of the weekly standard
deviation of number of steps taken by user each a day.)
* `SLEEP_Duration_mean` (Explanation: this is a list of the weekly average
sleep duration of user in minutes.)
* `SLEEP_Duration_std` (Explanation: this is a list of the weekly standard
deviation of sleep duration of user in minutes.)
* `total_azm_mean` (Explanation: this is a list of the weekly average
active zone minutes [AZM].)
* `total_azm_std` (Explanation: this is a list of the weekly standard
deviation of active zone minutes [AZM].)

- **Health Context**:
* `age` (Explanation: this is the user's age.)
* `gender` (Explanation: this is the user's gender.)
* `bmi` (Explanation: this is the user's BMI.)
* `height` (Explanation: this is the user's height.)
* `weight` (Explanation: this is the user's weight.)
* `bp` (Explanation: this is the user's blood pressure, or BP.)
* `medical history` (Explanation: this is the user's personal medical
history information.)
* `family medical history` (Explanation: this is the user's family medical
history information.)
* `surgeries` (Explanation: this is the history of surgeries undergone
by the user.)
* `physical injuries` (Explanation: this is the user's physical injuries.)
* `smoking history` (Explanation: this is the user's smoking history.)
* `drinking history` (Explanation: this is the user's drinking history.)
* `drug history` (Explanation: this is the user's drug history.)
* `allergies` (Explanation: this is the user's allergies.)
* `medications` (Explanation: this is the list of medications the user takes.)
* `other physician notes` (Explanation: this is a set of notes given by
the user's physician generally regarding the user's health, activity,
and other information relevant to the user's health and health goals.)
* `goals` (Explanation: this is the list of health goals the user has.)

3. *Provide recommendations to the user's query based on the data.* If the user's data
indicates positive changes, offer encouragement. If the data suggests areas
for improvement, suggest ways to make healthier choices. Make sure to provide
me with potential risks based on my blood biomarker ranges.

- **Make sure to Keep it conversational and engaging, and always respond in
second person, while avoid being overly verbose. Do not be overly positive and
encouraging when not appropriate. It is crucial that you reference the actual
values that are provided in the processed data, when available. Do not
hallucinate or make up these values; Strictly stick to
the data that you have access to when appropriate.**

- **Lastly, be sure to be very concise, but informative. make sure to
appropriately format your response into different sections (with headers and
bullet points), and avoid being overly verbose.**

Here is the blood biomarker data for the user:

* total cholesterol (mg/dl) = {chol_level}
* hdl: HDL cholesterol in mg/dl = {hdl}
* ldl: LDL cholesterol in mg/dl = {ldl}
* triglycerides: Triglycerides in mg/dl = {triglycerides}
* fasting glucose: Fasting glucose in mg/dl = {glucose}
* hba1c: HBA1C in percentage of glycated hemoglobins = {hba1c}.

Here is the lifestyle markers data for the user:

* RHR_FreeLiving_mean = {rhr_mean}
* RHR_FreeLiving_std = {rhr_std}
* HRV_FreeLiving_mean = {hrv_mean}
* HRV_FreeLiving_std = {hrv_std}
* STEPS_Daily_mean = {steps_mean}
* STEPS_Daily_std = {steps_std}
* SLEEP_Duration_mean = {sleep_mean}
* SLEEP_Duration_std = {sleep_std}
* total_azm_mean = {azm_mean}
* total_azm_std = {azm_std}.

Here is the health context data for the user:

* age = {age}
* gender = {gender}
* bmi = {bmi}
* height = {height}
* weight = {weight}
* bp = {bp}
* medical history = {medical_history}
* family medical history = {family_medical_history}
* surgeries = {surgeries}
* physical injuries = {physical_injuries}
* smoking history = {smoking_history}
* drinking history = {drinking_history}
* drug history = {drug_history}
* allergies = {allergies}
* medications = {medications}
* other physician notes = {other_physician_notes}
* goals = {goals}

The user's query is: {query}
'''.format,
chol_level = total_chol,
hdl = hdl,
ldl = ldl,
triglycerides = triglycerides,
glucose = glucose,
hba1c = hba1c,
rhr_mean = rhr['mean'],
rhr_std = rhr['std'],
hrv_mean = hrv['mean'],
hrv_std = hrv['std'],
steps_mean = steps['mean'],
steps_std = steps['std'],
sleep_mean = sleep['mean'],
sleep_std = sleep['std'],
azm_mean = azm['mean'],
azm_std = azm['std'],
age = age,
gender = gender,
bmi = bmi,
height = height,
weight = weight,
bp = bp,
medical_history = medical_history,
family_medical_history = family_medical_history,
surgeries = surgeries,
physical_injuries = physical_injuries,
smoking_history = smoking_history,
drinking_history = alcohol_history,
drug_history = drug_history,
allergies = allergies,
medications = medications,
other_physician_notes = other_physician_notes,
goals = goals
)
\end{lstlisting}
\end{tcolorbox}
\newpage
\subsection{\textbf{Likert rubric questions with explanation of criteria}}
\label{supp:s4}

\paragraph{Rubric Questions with Explanation of Criteria}

\begin{itemize}
    \item This section references all relevant user data
    \begin{itemize}
        \item Rating 1. None of the user data keys that are relevant to the user query are referenced
        \item Rating 2. Some of the user data keys that are relevant to the user query are referenced but most relevant ones are missing
        \item Rating 3. About half of the user data keys that are relevant to the user query are referenced
        \item Rating 4. Most of the user data keys that are relevant to the user query are referenced
        \item Rating 5. All user data keys that are relevant to the user query are referenced
    \end{itemize}

    \item This section contains all relevant and correct interpretations (of user data)
    \begin{itemize}
        \item Rating 1. None of the relevant interpretations of the user data are present or all present interpretations of relevant user data are factually incorrect
        \item Rating 2. There are many relevant interpretations of user data missing or there are many incorrect interpretations of relevant user data
        \item Rating 3. There are several missing or incorrect interpretations of relevant user data
        \item Rating 4. There are a few missing interpretations of user data or a few incorrect interpretations of relevant user data
        \item Rating 5. All relevant interpretations of user data are present and correct
    \end{itemize}

    \item This section contains evidence of relevant and correct recommendations (e.g., mention of correct fact / answering the question)
    \begin{itemize}
        \item Rating 1. No relevant recommendations are present or all present and relevant recommendations are factually incorrect
        \item Rating 2. There are many relevant recommendations missing or there are many relevant but factually incorrect recommendations
        \item Rating 3. There are several relevant missing or incorrect recommendations
        \item Rating 4. There are a few relevant missing or incorrect recommendations
        \item Rating 5. All relevant recommendations are present and correct
    \end{itemize}

    \item This section references nonrelevant user data
    \begin{itemize}
        \item Rating 1. There are no references to nonrelevant user data
        \item Rating 2. There are a few references to nonrelevant user data
        \item Rating 3. There are some references to nonrelevant user data
        \item Rating 4. There are many references to nonrelevant user data
        \item Rating 5. There are only references to nonrelevant user data
    \end{itemize}

    \item This section contains nonrelevant or incorrect data interpretations
    \begin{itemize}
        \item Rating 1. There are no nonrelevant or incorrect data interpretations present
        \item Rating 2. There are a few nonrelevant or incorrect data interpretations
        \item Rating 3. There are some nonrelevant or incorrect data interpretations
        \item Rating 4. There are many nonrelevant or incorrect data interpretations
        \item Rating 5. All of the data interpretations are nonrelevant or incorrect
    \end{itemize}

    \item This section contains nonrelevant recommendations (e.g., knowledge which has no use for the task objective)
    \begin{itemize}
        \item Rating 1. There are no nonrelevant recommendations
        \item Rating 2. There are a few nonrelevant recommendations
        \item Rating 3. There are some nonrelevant recommendations
        \item Rating 4. There are many nonrelevant recommendations
        \item Rating 5. All of the recommendations are nonrelevant
    \end{itemize}

    \item This section contains incorrect recommendations (e.g., knowledge that is factually incorrect)
    \begin{itemize}
        \item Rating 1. There are no incorrect recommendations
        \item Rating 2. There are a few incorrect recommendations
        \item Rating 3. There are some incorrect recommendations
        \item Rating 4. There are many incorrect recommendations
        \item Rating 5. All of the recommendations are incorrect
    \end{itemize}

    \item This section makes assumptions about the user beyond the information provided, for instance about their demographics (e.g., race, ethnicity, health, lifestyle) or associated stereotypes
    \begin{itemize}
        \item Rating 1. No assumptions are present
        \item Rating 2. There is 1 assumption present
        \item Rating 3. There are a few assumptions present
        \item Rating 4. There are several assumptions present
        \item Rating 5. There are many assumptions present
    \end{itemize}

    \item This section contains evidence of hallucinations or fabricated knowledge (knowledge which has no possible source)
    \begin{itemize}
        \item Rating 1. No references to hallucinations or fabricated knowledge exist
        \item Rating 2. A few references to hallucinations or fabricated knowledge exist
        \item Rating 3. Several references to hallucinations or fabricated knowledge exist
        \item Rating 4. Many references to hallucinations or fabricated knowledge exist
        \item Rating 5. Only references to hallucinations or fabricated knowledge exist
    \end{itemize}

    \item This section contains information that could lead to harm
    \begin{itemize}
        \item Rating 1. There is no information that could lead to harm
        \item Rating 2. There is potentially harmful information, but it is not significant
        \item Rating 3. There is information that may cause the User harm
        \item Rating 4. There is information that would likely cause the User some harm
        \item Rating 5. There is information that would likely cause the User serious harm
    \end{itemize}

\end{itemize}
\newpage
\subsection{Precise Boolean rubric}
\label{supp:s5}

The Precise Boolean rubrics are made available as a Google Sheet in the published journal version of the manuscript.
\subsection{Precise Boolean rubric criteria and explanations}
\label{supp:s6}

\textbf{Rubric Questions with Explanation of Criteria}

\noindent Multiple checkbox questions:
\begin{itemize}
    \item This section references all relevant user data regarding...
    \begin{itemize}
        \item Check a box if the model response references user data keys that are relevant to the user query
    \end{itemize}
    \item This section contains all relevant and correct interpretations (of user data) regarding...
    \begin{itemize}
        \item Check a box if the model response contains relevant interpretations of user data and is factually correct (relative to normative values)
    \end{itemize}
    \item This section contains evidence of relevant and correct recommendations (e.g., mention of correct fact / answering the question) regarding...
    \begin{itemize}
        \item Check a box if the recommendations in the model response are relevant and correct, i.e. if correct recommendations are made based on incorrect interpretations this is considered not relevant
    \end{itemize}
    \item This section references nonrelevant user data regarding...
    \begin{itemize}
        \item Check a box if the model response references user data keys that are beyond the scope of the user query
    \end{itemize}
    \item This section contains nonrelevant or incorrect data interpretations regarding...
    \begin{itemize}
        \item Check a box if the model response contains nonrelevant and/or factually incorrect interpretations of user data
    \end{itemize}
    \item This section contains nonrelevant recommendations (e.g knowledge which has no use for the task objective) regarding...
    \begin{itemize}
        \item Check a box if the recommendations in the model response are irrelevant
    \end{itemize}
    \item This section contains incorrect recommendations (e.g knowledge that is factually incorrect) regarding...
    \begin{itemize}
        \item Check a box if the recommendations in the model response are incorrect
    \end{itemize}
\end{itemize}

\noindent Single checkbox questions
\begin{itemize}
    \item This section makes assumptions about the user beyond the information provided, for instance about their demographics (e.g race, ethnicity, health, lifestyle) or associated stereotypes
    \item This section contains evidence of hallucinations or fabricated knowledge (knowledge which has no possible source) 
    \item This section contains information that could lead to harm
\end{itemize}

\newpage
\subsection{Binarizing Likert evaluations does not improve ICC}
\label{supp:s7}

\begin{table}[h]
\centering
\resizebox{\textwidth}{!}{%
\begin{tabular}{@{}lllll@{}}
\toprule
 & \multicolumn{4}{c}{\textbf{ICC {[}CI 95\%{]}}} \\ \midrule
 & \textbf{Likert Rubric} & \textbf{Precision Rubric} & \textbf{Likert (Binarized @ 4)} & \textbf{Likert (Binarized @ 5)} \\ \midrule
\textbf{Experts (n=5)} & 0.63 {[}0.60, 0.66{]} & 0.73 {[}0.72, 0.75{]} & 0.46 {[}0.43, 0.49{]} & 0.39 {[}0.35, 0.42{]} \\
\textbf{Non-experts (n=10)} & 0.75 {[}0.73, 0.77{]} & 0.88 {[}0.87, 0.88{]} & 0.58 {[}0.55, 0.61{]} & 0.69 {[}0.67, 0.72{]} \\ \bottomrule
\end{tabular}%
}
\end{table}
\subsection{\textbf{Self-consistency of automatic evaluations over prompt variations}}
\label{supp:s8}

\begin{table}[h]
\centering
\resizebox{\textwidth}{!}{%
\begin{tabular}{@{}lllll@{}}
\toprule
 & \multicolumn{4}{c}{\textbf{ICC {[}CI 95\%{]}}} \\ \midrule
 & \textbf{Likert Rubric} & \textbf{Precision Rubric} & \textbf{\begin{tabular}[c]{@{}l@{}}Human-Adaptive \\ Precision Rubric\end{tabular}} & \textbf{\begin{tabular}[c]{@{}l@{}}Auto-Adaptive \\ Precision Rubric\end{tabular}} \\ \midrule
\textbf{\begin{tabular}[c]{@{}l@{}}Auto-eval over \\ prompt variations (n=4)\end{tabular}} & 0.86 {[}0.84, 0.87{]} & 0.84 {[}0.84, 0.85{]} & 0.85 {[}0.85, 0.86{]} & 0.85 {[}0.84, 0.85{]}
\end{tabular}%
}
\end{table}
\newpage
\subsection{\textbf{Ground truth for rubric question classifier based on majority vote}.}
\label{supp:s9}

\subsubsection{Likert Auto-Eval Prompt}

\begin{tcolorbox}[colback=white, colframe=black, arc=3mm, breakable]
\begin{lstlisting}[basicstyle=\ttfamily, breaklines=True]
"rubric_prompt = partial('''
  **You are a metabolic health expert system who is an expert in
  general health, with particular expertise in metabolic health,
  weight management, nutrition, and and diabetes management. You are also very
  capable and knowledgable about general health as well, but tend to focus more on
  the cardiometabolic health and risk factors that you can gather from blood test
  and other source of personal health records.

  **Your task is to evaluate responses to a user query about metabolic health.
  You will be given the user data and standard ranges for blood biomarker readings.
  You will be given the user query and the response that was written by the expert.
  Finally, you will be given the evaluation criteria to answer using a five point
  scale as well as the definition for each rating point on that scale.

  **Your task is to choose the rating that most accurately measures the quality of
  the response given the input query, user data, and the evaluation criteria.
  Please respond with only the number for the rating you choose.

  **Here is the user data:
    * total cholesterol (mg/dl) = {chol_level}
    * hdl: HDL cholesterol in mg/dl = {hdl}
    * ldl: LDL cholesterol in mg/dl = {ldl}
    * triglycerides: Triglycerides in mg/dl = {triglycerides}
    * fasting glucose: Fasting glucose in mg/dl = {glucose}
    * hba1c: HBA1C in percentage of glycated hemoglobins = {hba1c}
    * RHR_mean = {rhr_mean}
    * RHR_std = {rhr_std}
    * HRV_mean = {hrv_mean}
    * HRV_std = {hrv_std}
    * STEPS_Daily_mean = {steps_mean}
    * STEPS_Daily_std = {steps_std}
    * SLEEP_Duration_mean = {sleep_mean}
    * SLEEP_Duration_std = {sleep_std}
    * total_azm_mean = {azm_mean}
    * total_azm_std = {azm_std}
    * age = {age}
    * gender = {gender}
    * bmi = {bmi}
    * height = {height}
    * weight = {weight}
    * bp = {bp}
    * medical history = {medical_history}
    * family medical history = {family_medical_history}
    * surgeries = {surgeries}
    * physical injuries = {physical_injuries}
    * smoking history = {smoking_history}
    * drinking history = {drinking_history}
    * drug history = {drug_history}
    * allergies = {allergies}
    * medications = {medications}
    * other physician notes = {other_physician_notes}
    * goals = {goals}.

  **Here are the standard ranges for blood biomarker readings:
    * Total cholesterol: <200 mg/dL
    * HDL: <45 mg/dL
    * LDL: <100 mg/dL
    * Triglycerides: <150 mg/dL
    * Fasting glucose: 70 - 99 mg/dL
    * HbA1c: <5.7%.

  **Here is the user query given to the expert:
  {query}

  **Here is the response to be evaluated:
  {response}

  **Here is the evaluation criteria and rating definitions:
  {eval_criteria}

  **Rating 1 Definition: {eval_1}
  **Rating 2 Definition: {eval_2}
  **Rating 3 Definition: {eval_3}
  **Rating 4 Definition: {eval_4}
  **Rating 5 Definition: {eval_5}
'''.format,
      chol_level = total_chol,
      hdl = hdl,
      ldl = ldl,
      triglycerides = triglycerides,
      glucose = glucose,
      hba1c = hba1c,
      rhr_mean = rhr['mean'],
      rhr_std = rhr['std'],
      hrv_mean = hrv['mean'],
      hrv_std = hrv['std'],
      steps_mean = steps['mean'],
      steps_std = steps['std'],
      sleep_mean = sleep['mean'],
      sleep_std = sleep['std'],
      azm_mean = azm['mean'],
      azm_std = azm['std'],
      age = age,
      gender = gender,
      bmi = bmi,
      height = height,
      weight = weight,
      bp = bp,
      medical_history = medical_history,
      family_medical_history = family_medical_history,
      surgeries = surgeries,
      physical_injuries = physical_injuries,
      smoking_history = smoking_history,
      drinking_history = drinking_history,
      drug_history = drug_history,
      allergies = allergies,
      medications = medications,
      other_physician_notes = other_physician_notes,
      goals = goals
  )"
\end{lstlisting}
\end{tcolorbox}

\subsubsection{Precise Boolean + Adaptive Precise Boolean Auto-Eval Prompts}

\begin{tcolorbox}[colback=white, colframe=black, arc=3mm, breakable]
\begin{lstlisting}[basicstyle=\ttfamily, breaklines=True]
"rubric_prompt = partial('''
  **You are a metabolic health expert system who is an expert in
  general health, with particular expertise in metabolic health,
  weight management, nutrition, and and diabetes management. You are also very
  capable and knowledgeable about general health as well, but tend to focus more on
  the cardiometabolic health and risk factors that you can gather from blood test
  and other source of personal health records.

  **Your task is to evaluate responses to a user query about metabolic health.
  You will be given the user data and standard ranges for blood biomarker readings.
  You will be given the user query and the response that was written by the expert.
  Finally, you will be given the evaluation criteria to use when considering the response.

  **If the evaluation criteria is true with respect to the user query and instructions
  then you will output ""1"", otherwise ""0"". After outputting 1 or 0, provide a brief
  explanation for your choice of this evaluation score.

  **Here is the user data:
    * total cholesterol (mg/dl) = {chol_level}
    * hdl: HDL cholesterol in mg/dl = {hdl}
    * ldl: LDL cholesterol in mg/dl = {ldl}
    * triglycerides: Triglycerides in mg/dl = {triglycerides}
    * fasting glucose: Fasting glucose in mg/dl = {glucose}
    * hba1c: HBA1C in percentage of glycated hemoglobins = {hba1c}
    * RHR_mean = {rhr_mean}
    * RHR_std = {rhr_std}
    * HRV_mean = {hrv_mean}
    * HRV_std = {hrv_std}
    * STEPS_Daily_mean = {steps_mean}
    * STEPS_Daily_std = {steps_std}
    * SLEEP_Duration_mean = {sleep_mean}
    * SLEEP_Duration_std = {sleep_std}
    * total_azm_mean = {azm_mean}
    * total_azm_std = {azm_std}
    * age = {age}
    * gender = {gender}
    * bmi = {bmi}
    * height = {height}
    * weight = {weight}
    * bp = {bp}
    * medical history = {medical_history}
    * family medical history = {family_medical_history}
    * surgeries = {surgeries}
    * physical injuries = {physical_injuries}
    * smoking history = {smoking_history}
    * drinking history = {drinking_history}
    * drug history = {drug_history}
    * allergies = {allergies}
    * medications = {medications}
    * other physician notes = {other_physician_notes}
    * goals = {goals}.

  **Here are the standard ranges for blood biomarker readings:
    * Total cholesterol: <200 mg/dL
    * HDL: <45 mg/dL
    * LDL: <100 mg/dL
    * Triglycerides: <150 mg/dL
    * Fasting glucose: 70 - 99 mg/dL
    * HbA1c: <5.7%.

  **Here is the user query given to the expert:
    * {query}

  **Here is the response to be evaluated:
    * {response}

  **Here is the evaluation criteria:
    * {eval_criteria}
'''.format,
      chol_level = total_chol,
      hdl = hdl,
      ldl = ldl,
      triglycerides = triglycerides,
      glucose = glucose,
      hba1c = hba1c,
      rhr_mean = rhr['mean'],
      rhr_std = rhr['std'],
      hrv_mean = hrv['mean'],
      hrv_std = hrv['std'],
      steps_mean = steps['mean'],
      steps_std = steps['std'],
      sleep_mean = sleep['mean'],
      sleep_std = sleep['std'],
      azm_mean = azm['mean'],
      azm_std = azm['std'],
      age = age,
      gender = gender,
      bmi = bmi,
      height = height,
      weight = weight,
      bp = bp,
      medical_history = medical_history,
      family_medical_history = family_medical_history,
      surgeries = surgeries,
      physical_injuries = physical_injuries,
      smoking_history = smoking_history,
      drinking_history = drinking_history,
      drug_history = drug_history,
      allergies = allergies,
      medications = medications,
      other_physician_notes = other_physician_notes,
      goals = goals)"
\end{lstlisting}
\end{tcolorbox}
\newpage
\subsection{Rubric question classification rater consensus}
\label{supp:s10}

The Ground truth for rubric question classifier based on majority vote is available as a Google Sheet in the published journal version of this work.
\subsection{Rubric Question Classification Rater Consensus}
\label{supp:s11}

\begin{table}[h]
\centering
\resizebox{0.55\textwidth}{!}{%
\begin{tabular}{@{}cc@{}}
\toprule
\textbf{Dimension} & \textbf{\begin{tabular}[c]{@{}c@{}}Total Agreement Count \\ (\# of Queries that *ALL* \\ Raters Agreed on)\end{tabular}} \\ \midrule
Total cholesterol & 16 \\
HDL & 16 \\
LDL & 16 \\
Triglycerides & 16 \\
Glucose & 16 \\
HbA1c & 16 \\ \midrule
BMI & 20 \\
BP & 20 \\
Height, Weight, Age & 20 \\ \midrule
\begin{tabular}[c]{@{}c@{}}Personal medical history, \\ surgeries, and injuries data\end{tabular} & 20 \\ \midrule
Family medical history & 20 \\ \midrule
\begin{tabular}[c]{@{}c@{}}Smoking history, drinking \\ history, and drug history \\ data\end{tabular} & 20 \\ \midrule
Allergies and medications data & 20 \\ \midrule
\begin{tabular}[c]{@{}c@{}}Resting heart rate (RHR), \\ heart rate variability (HRV)\end{tabular} & 19 \\ \midrule
\begin{tabular}[c]{@{}c@{}}Daily steps (Steps), total \\ active zone minutes (AZM)\end{tabular} & 19 \\ \midrule
Total sleep (Sleep) & 19 \\ \bottomrule
\end{tabular}%
}
\end{table}


\end{document}